\pdfoutput=1

\documentclass[11pt]{article}

\usepackage[final]{acl}

\usepackage{times}
\usepackage{latexsym}
\usepackage{amsfonts}

\usepackage[T1]{fontenc}

\usepackage[utf8]{inputenc}

\usepackage{microtype}

\usepackage{inconsolata}

\usepackage{graphicx}
\usepackage{amsmath}
\usepackage{cleveref}
\usepackage{todonotes}
\usepackage{booktabs}
\usepackage{xltabular}
\usepackage{comment}
\usepackage{subcaption}

\usepackage{hyperref}

\title{Personalized Benchmarking: Evaluating LLMs by Individual Preferences}

\author{
  \textbf{Cristina G\^arbacea$^*$},
  \textbf{Heran Wang$^*$},
  \textbf{Chenhao Tan}
\\
{\{garbacea, heranwang, chenhao\}@uchicago.edu}
\\
 University of Chicago
}

\begin{document}
\maketitle

{\def\thefootnote{*}\footnotetext{Equal contribution.}}

\begin{abstract}

With the rise in capabilities of large language models (LLMs) and their deployment in real-world tasks, evaluating LLM alignment with human preferences has become an important challenge. Current benchmarks average preferences across all users to compute aggregate ratings, overlooking individual user preferences when establishing model rankings. Since users have varying preferences in different contexts, we call for personalized LLM benchmarks that rank models according to individual needs. We compute personalized model rankings using ELO ratings and Bradley-Terry coefficients for 115 active Chatbot Arena users and analyze how user query characteristics (topics and writing style) relate to LLM ranking variations. We demonstrate that individual rankings of LLM models diverge dramatically from aggregate LLM rankings, with Bradley-Terry correlations averaging only $\rho = 0.04$ (57\% of users show near-zero or negative correlation) and ELO ratings showing moderate correlation ($\rho = 0.43$). Through topic modeling and style analysis, we find users exhibit substantial heterogeneity in topical interests and communication styles, influencing their model preferences. We further show that a compact combination of topic and style features provides a useful feature space for predicting user-specific model rankings. Our results provide strong quantitative evidence that aggregate benchmarks fail to capture individual preferences for most users, and highlight the importance of developing personalized benchmarks that rank LLM models according to individual user preferences.

\end{abstract}

\section{Introduction}
\label{sec::introduction}

Large language models (LLMs) have demonstrated remarkable capabilities across a wide range of tasks, from creative writing and code generation to question answering and complex reasoning. As these models become increasingly integrated into everyday applications, the need for robust evaluation methodologies has grown correspondingly important. Traditional benchmarks for LLMs have primarily focused on aggregate performance metrics, measuring model capabilities on standardized tasks or averaging human preferences across diverse user populations to establish the best performing model.

However, this one-size-fits-all approach to LLM evaluation overlooks a fundamental aspect of real-world deployment: individual users have diverse preferences, needs, and use cases. A model that excels at technical documentation may not be ideal for creative storytelling. Similarly, users may have different preferences regarding response verbosity, formality, or the inclusion of explanations versus direct answers. These individual differences are averaged away in current benchmarking approaches, potentially leading to suboptimal model recommendations for specific users or use cases.

For example, consider a software developer who prefers concise, technically precise responses with code examples, versus a creative writer who values elaborate, imaginative responses with rich vocabulary. Current benchmarks would rank models based on averaged preferences across both users, potentially recommending a model that is mediocre for both rather than identifying which model best serves each individual's needs. This aggregation problem becomes more pronounced as LLMs are deployed to increasingly diverse user populations with heterogeneous preferences.

In this work, we address this gap by introducing the concept of \textbf{personalized benchmarking} for LLM alignment evaluation based on individual user preferences. Our key contributions are:

\begin{itemize}
\item We formalize the problem of personalized LLM evaluation and compute per-user model rankings using ELO ratings and Bradley-Terry coefficients for active Chatbot Arena users;

\item We provide quantitative evidence that individual rankings diverge substantially from aggregate rankings for the majority of users, under both ELO and Bradley-Terry rating systems;

\item We characterize user heterogeneity along two interpretable dimensions (query topics and writing style) identifying stable, systematic differences across users;

\item We demonstrate that combining a compact topic profile with interpretable style features provides a meaningful feature space for predicting user-specific LLM model rankings.

\end{itemize}

Our findings reveal that users demonstrate substantial heterogeneity in both their query patterns and model preferences. By acknowledging and quantifying this preference heterogeneity, we can develop more nuanced systems that match users to appropriate models and ultimately improve user satisfaction in LLM-powered applications.

\section{Related Work}
\label{sec::related_work}

\paragraph{LLM Benchmarking and Evaluation} %
Chatbot Arena \cite{zheng2023judging, chiang2024chatbot} introduced a crowdsourced platform where users compare model responses head-to-head, aggregating preferences into ELO ratings. Similarly, Alpaca Eval \cite{alpaca_eval} and MT-Bench \cite{bai2024mt} evaluate models on instruction-following capabilities using both automated and human evaluation. Arena Hard \cite{li2025crowdsourced} extended this paradigm with challenging user queries designed to differentiate between top performing models.

However, all these benchmarks compute aggregate rankings by averaging across all evaluations, implicitly assuming preference homogeneity across users. Our work challenges this assumption and demonstrates the value of personalized evaluation.

\paragraph{Preference Learning and Ranking Systems} The Bradley-Terry (BT) model \cite{bradley1952rank} provides a probabilistic framework for pairwise comparisons, estimating the probability that one item is preferred over another. ELO \citep{elo1967proposed} rating systems, originally developed for chess, have been adapted for LLM evaluation.

Reinforcement learning from human feedback (RLHF) \cite{ouyang2022training, lambert2025reinforcement} has popularized preference learning for LLM alignment. Models are fine-tuned to maximize expected human preferences, typically estimated from crowdsourced pairwise comparisons. However, this work generally treats human preferences as a single aggregate signal rather than modeling individual differences.

\paragraph{LLM Personalization}
HyPerAlign \cite{garbacea2025hyperalign}  focuses on interpretable personalized LLM alignment via hypothesis generation, demonstrating that individual user preferences can be captured and modeled. Our work complements these efforts by providing a framework for measuring how well different models align with individual user preferences through query analysis.

Research in human-AI interaction has highlighted the importance of understanding user needs and preferences when deploying AI systems \cite{ehsan2021expanding}. Users have diverse expectations for AI assistants, varying in their tolerance for errors, preference for explanation, and desired level of autonomy. 
However, most evaluation benchmarks have not systematically incorporated these insights; instead they rely on aggregate metrics.

\section{Methodology}
\label{sec::methodology}

\subsection{Problem Formulation}

Let $\mathcal{M} = \{m_1, m_2, \ldots, m_M\}$ denote a set of LLM models to be evaluated, and $\mathcal{U} = \{u_1, u_2, \ldots, u_U\}$ denote a set of users. Each user $u_i$ interacts with models by submitting queries and evaluating responses through pairwise comparisons. For a given query $q$, two models $m_a$ and $m_b$ generate responses $r_a$ and $r_b$, and the user indicates their preference as one of the following options: $m_a \succ m_b$, $m_b \succ m_a$, or tie. Traditional benchmarking aggregates all comparisons across users to compute a single global ranking $R_{global}: \mathcal{M} \rightarrow \mathbb{R}$. In contrast, \textit{personalized benchmarking computes individual rankings $R_{u_i}: \mathcal{M} \rightarrow \mathbb{R}$ for each user $u_i$}, therefore capturing user-specific preferences. 

Our goal is threefold: \textit{i)} compute reliable personalized rankings for individual users and determine to what extent they correlate with aggregate rankings of LLM models, \textit{ii)} analyze user query patterns in terms of topics and writing styles, \textit{iii)} identify dimensions of user query heterogeneity that may explain preference variations.

\subsection{Rating Systems for Personalized Rankings}

We employ two established rating systems to compute personalized model rankings: ELO ratings and the Bradley-Terry model.

\paragraph{ELO Rating System} Originally developed for chess, the ELO rating system maintains a numerical rating for each model that is updated after each pairwise model comparison. For a given user $u$, we maintain model ratings $\{ELO_u(m) | m \in \mathcal{M}\}$. When the user $u$ prefers model $m_a$ over model $m_b$, the ratings are updated as follows:

\begin{align}
E_a &= \frac{1}{1 + 10^{(ELO_u(m_b) - ELO_u(m_a))/400}} \\
E_b &= 1 - E_a \\
ELO_u(m_a) &\leftarrow ELO_u(m_a) + K(1 - E_a) \\
ELO_u(m_b) &\leftarrow ELO_u(m_b) + K(0 - E_b)
\end{align}
where $E_a$ represents the expected score for model $m_a$ and $K$ is a constant determining the update magnitude (we use $K=32$). For ties, both models receive a score of 0.5.

\paragraph{Bradley-Terry Model}

The Bradley-Terry model provides a probabilistic framework for pairwise comparisons. It assumes that each model $m$ has an underlying strength parameter $\beta_m$, and the probability that user $u$ prefers model $m_a$ over $m_b$ is computed as:

\begin{equation}
P(m_a \succ_u m_b) = \frac{\beta_{u,m_a}}{\beta_{u,m_a} + \beta_{u,m_b}}
\end{equation}

User-specific parameters $\{\beta_{u,m} | m \in \mathcal{M}\}$ are estimated by maximum likelihood estimation. Given user $u$'s comparison data $\mathcal{D}_u = \{(m_a^{(i)}, m_b^{(i)}, y^{(i)})\}$ where $y^{(i)} \in \{0, 1\}$ indicates whether $m_a$ was preferred, the log-likelihood used to estimate the $\beta_{u,m}$ parameters is modeled as:

\begin{equation}
\begin{split}
\mathcal{L}(\{\beta_{u,m}\}) &= \sum_i y^{(i)} \log \beta_{u,m_a^{(i)}} \\&- \log(\beta_{u,m_a^{(i)}} + \beta_{u,m_b^{(i)}})
\end{split}
\end{equation}

\subsection{User Query Analysis}

To understand what drives potential ranking variations among users, we analyze user queries along two primary dimensions, namely topic and style.

\paragraph{Topic Modeling}
We use FastTopic \cite{wu2024fastopic}, a dual semantic-relation reconstruction based neural topic model, to construct topic profiles for each user. FastTopic employs an embedding transport plan objective to better align document and topic representations, which helps reduce repetitive and low-quality topics while improving topic distinctiveness. We first form a global prompt corpus by aggregating queries from all users, $\mathcal{Q} = \bigcup_{i} Q_{u_i}$, where each user $u_i$ has a set of queries $Q_{u_i} = \{q_1, q_2, \ldots, q_{n_i}\}$. We then train FastTopic on $\mathcal{Q}$ to learn a shared set of $K$ latent topics, so that all users are represented in the same topic space. For each query $q_j \in Q_{u_i}$, FastTopic produces a topic-mixture vector \[ \boldsymbol{\theta}_{ij} = (\theta_{ij1}, \theta_{ij2}, \ldots, \theta_{ijK}), \qquad \sum_{k=1}^{K} \theta_{ijk} = 1, \] where $\theta_{ijk}$ denotes the proportion of topic $k$ in query $q_j$. We define the topic profile of user $u_i$ as the mean topic mixture across that user’s queries: \[ \mathbf{t}_{u_i} = \frac{1}{n_i} \sum_{j=1}^{n_i} \boldsymbol{\theta}_{ij}. \] Thus, $\mathbf{t}_{u_i} \in \mathbb{R}^{K}$ is a $K$-dimensional representation of the user’s topical preferences, with each dimension corresponding to the user’s average affinity for one global topic. This setup allows different users to be compared directly as different mixtures over the same shared topic basis.

\paragraph{Style Analysis} We employ a hybrid approach for style extraction, combining quantitative embeddings with qualitative characterization. 

\textit{LISA-based Style Modeling} We use LISA \cite{patel2023learning, malik2024empirical} to construct interpretable style embeddings. LISA embeds queries into a 768-dimensional style vector space, where each dimension corresponds to a specific stylistic attribute. We first embed all queries using LISA and extract the top 20 dimensions for each query based on their activation strength. These style attribute indices are then treated as documents and used to train an LDA model with 6 style topics. To make the resulting style topics interpretable, we generate  labels for each topic using Claude Sonnet 4.5 \cite{anthropic2025claude45sonnet} and GPT-5 \cite{openai2025gpt5} based on the  style attributes. Finally, we compute mean style vectors for each user by averaging their query LDA embeddings, yielding a compact representation of each user's writing style.

\textit{HypoGeniC Style Characterization} To complement the quantitative LISA analysis, we employ a qualitative approach following the HyPerAlign framework \cite{garbacea2025hyperalign}. We use HypoGeniC \cite{zhou2024hypothesis} on top of  Meta-Llama-3-8B-Instruct \cite{grattafiori2024llama} to generate natural language hypotheses about each user's writing style based on their query texts. This approach produces rich, interpretable characterizations that capture nuances in tone, structure, and rhetorical patterns that may not be fully represented in the quantitative LISA style embeddings.

\subsection{User Segmentation}

We cluster users based on their style patterns derived from LISA interpretable style embeddings. This reveals whether certain user types exhibit consistent writing characteristics that may correlate with different model preferences. User clustering employs k-means on the 768-dimensional  style feature vectors; the number of clusters is selected via silhouette analysis \cite{rousseeuw1987silhouettes}, resulting in $k=3$ user clusters. 

\section{Experimental Setup}

\subsection{Datasets}

We leverage the publicly available Chatbot Arena Conversations \cite{zheng2023judging}\footnote{\url{https://huggingface.co/datasets/lmsys/chatbot_arena_conversations}} dataset, which contains user queries and pairwise preference votes expressed by diverse users over different LLM models. We filter users to focus the analysis only on those users who have participated in at least 25 battles (i.e., pairwise preference votes across two LLM models). This results in a subset of 115 active users out of the total 13,383 users in the dataset, with an average number of 59 unique queries per user (min: 25, max: 380, std: 56.61).

\subsection{Implementation Details}

\paragraph{Personalized Rankings} We use the publicly available code released by Chatbot Arena\footnote{\url{https://colab.research.google.com/drive/1XAO_wTTaruygWgcZmbeII-RGJ2PPfxqX?usp=sharing}} authors to compute personalized ELO ratings and Bradley-Terry coefficients for each user based on their preference votes. This allows us to establish individual model rankings that reflect each user's specific preferences rather than aggregate preferences across all users. To quantify the degree to which individual user preferences diverge from aggregate rankings, we compute Spearman \cite{spearman1904proof} rank correlations between global model rankings and each individual user's personalized ELO and Bradley-Terry rankings. Critically, we restrict the analysis only to models that each user has actually evaluated, and exclude models for which the respective user has not participated in any  battles (identified by default ELO scores of 1000, or NaN values in Bradley-Terry). This ensures correlations reflect genuine preference alignment rather than artifacts from unobserved model-user pairs.

\paragraph{Topic Modeling} We train a single global FastTopic model on the pooled prompt corpus from all users rather than fitting separate topic models for each user. Since all users are placed in a shared topic space, the resulting user-level topic proportions are directly comparable; in addition, this also provides a common set of topic features for the subsequent regression analysis. The topic profile for each user is constructed by averaging user-prompts topic distributions, which preserves soft topic membership and yields a stable fixed-length representation of the user’s overall topic preferences.

\paragraph{Style Analysis} We generate 768-dimensional style embeddings using LISA on top of all-MiniLM-L6-v2 sentence transformer \cite{reimers2019sentence}\footnote{\url{https://huggingface.co/sentence-transformers/all-MiniLM-L6-v2}}. To characterize the writing style of each user, we follow \citep{malik2024empirical}  and train an LDA model using the extracted LISA embeddings to derive 6 coarse-grained style attributes. 

For qualitative style characterization, we run HypoGeniC on Meta-Llama-3-8B-Instruct; the model is configured to generate 10 hypotheses for each user describing characteristics of their writing style in natural language. We do not apply any pre-processing for user queries, but instead retain all punctuation, capitalization and structural elements to preserve the maximum level of stylistic information that uniquely identifies each user.

\paragraph{Regression} We concatenate each user’s FastTopic topic profile with their LISA style embedding vector for training a regression model. For each user $u_i$, the input to the model is: \[ \mathbf{x}_{u_i} = [\mathbf{t}_{u_i}; \mathbf{s}_{u_i}] \in \mathbb{R}^{778}, \] where $\mathbf{t}_{u_i} \in \mathbb{R}^{10}$ is their FastTopic profile, and $\mathbf{s}_{u_i} \in \mathbb{R}^{768}$ is their interpretable style representation. The regression target is a 20-dimensional rating vector $\mathbf{y}_{u_i} \in \mathbb{R}^{20}$ corresponding to either ELO or Bradley-Terry (BT) model preference scores. We split users into a training and validation set, and standardize training features $\boldsymbol{x}$ and targets $\boldsymbol{y}$ to ensure comparability across regression settings.

For predicting ELO scores, we train an ensemble of 50 multilayer perceptrons with early stopping. For Bradley-Terry, we use a single dense multilayer perceptron with dropout. Both use Adam optimization; full details are in Appendix \Cref{appendix:regression_model}.

\section{Results}
\label{sec::results}

\subsection{Correlation Between Individual and Global Rankings of LLM Models}

\begin{figure*}[t!]
    \centering
    \includegraphics[width=\textwidth]{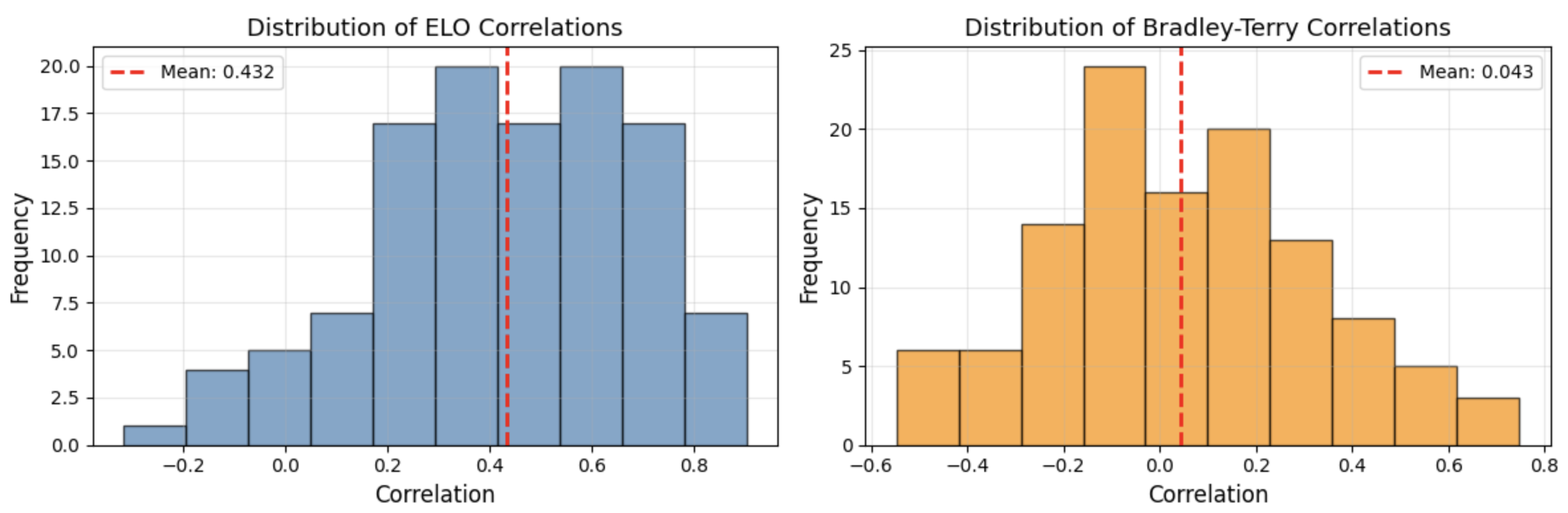}
        \caption{Spearman rank correlation between individual and global LLM rankings computed using ELO and Bradley-Terry rating systems. ELO shows moderate alignment  (mean $\rho= 0.43$), while Bradley-Terry reveals near-zero correlation ($\rho = 0.04$), with $57\%$ of users statistically indistinguishable from random model ordering.}
\label{fig:correlation_distribution}
\end{figure*}

In \Cref{fig:correlation_distribution} we present Spearman rank correlation results between global and individual rankings of LLM models computed using ELO and Bradley-Terry rating systems for the 115 active users in the Chatbot Arena dataset.

ELO-based individual rankings show only moderate alignment with the global consensus, with a mean Spearman correlation of $\rho= 0.432$ (SD = 0.257, median = 0.442, range: -0.317 to 0.903). Approximately 70\% of users (80 out of 115) exhibit correlations below 0.50, and several users show weakly negative correlations, suggesting their model preferences run counter to the aggregate ranking. Only a small fraction of users achieve correlations above 0.80, and the high standard deviation reflects substantial heterogeneity across the population. While ELO rankings do maintain a positive mean correlation, indicating some broad tendency for individual preferences to track the global order, the wide spread of values demonstrates that this aggregate signal is a poor proxy for the preferences of the majority of users, providing further quantitative motivation for personalized benchmarking approaches.

Bradley-Terry rankings reveal a far more striking divergence from the global consensus. Across the 115 users, the mean Spearman correlation is $\rho = 0.043$ (SD = 0.283, median = 0.011, range: -0.547 to 0.747), indicating that the typical user's personalized Bradley-Terry ranking bears essentially no relationship to the aggregate. Strikingly, 57\% of users (66 out of 115) exhibit correlations below 0.1, which suggests that for the majority of users, individual and global rankings are statistically indistinguishable from a random ordering of models. A substantial fraction of users exhibit negative correlations, with some reaching as low as -0.547 -- for these individuals the global ranking is actively misleading as a guide to model selection. Overall, 94\% of users (108 out of 115) show correlations below 0.5, while only a handful achieve correlations above 0.5, and even these remain modest in absolute terms. The high standard deviation further reflects that users are not clustered around any consistent degree of alignment with the aggregate: \textit{preference divergence is both widespread and heterogeneous}. 

This methodological discrepancy between ELO and Bradley-Terry is itself revealing. The Bradley-Terry model, which estimates win probabilities from pairwise comparisons through maximum likelihood estimation, appears substantially more sensitive to individual preference variations than the ELO system's incremental rating updates. This difference suggests that the choice of ranking algorithm affects not only the absolute rankings themselves, but also the apparent degree of personalization captured in the data. The Bradley-Terry method's probabilistic framework may better capture nuanced preference patterns that ELO's simpler update mechanism averages away.

To assess the reliability of these findings, we compute bootstrap confidence intervals ($B = 10,000$) and Wilcoxon signed-rank significance tests for both rating systems. ELO rankings show a mean Spearman correlation of $\rho = 0.432$ ($95\%$ CI $[0.387,0.478]$; $p < 10^{-19}$ against a null of zero), confirming that moderate alignment with the global consensus is a reliable population-level finding. In contrast, Bradley-Terry rankings yield a mean of $\rho = 0.043$ ($95\%$ CI $[-0.009,0.095]$; $p=0.165$), which is not significantly different from zero; this means that at the population level, individual BT rankings are statistically indistinguishable from a random ordering of models. This divergence between the two methods is itself highly significant (paired Wilcoxon $p < 10^{-13}$, confirming that ELO and BT capture fundamentally different signals rather than varying around a common truth. At the individual level, 42.6\% of users show statistically significant ELO correlations ($p < 0.05$), compared to only $9.6\%$ under Bradley-Terry which is barely above the 5\% false positive rate expected by chance. This further confirms that for the vast majority of users, personalized BT rankings bear no meaningful relationship to the global consensus.

Taken together, these results provide strong quantitative evidence that global LLM rankings diverge substantially from personalized rankings for the vast majority of users and obscure substantial individual preference heterogeneity. Users do not simply have slightly different preferences around a common ordering; rather, many \textbf{users have fundamentally different model preferences} that bear little resemblance to the global consensus. Relying on aggregate rankings to guide model selection for these users would likely result in suboptimal recommendations. Our findings also demonstrate that a single aggregate ordering of models is insufficient to reflect the diversity of individual preferences, and motivate the need for personalized benchmarking approaches.

\subsection{Topic Heterogeneity Across Users}

\begin{figure}[t]
\begin{center}
\includegraphics[width=\columnwidth]{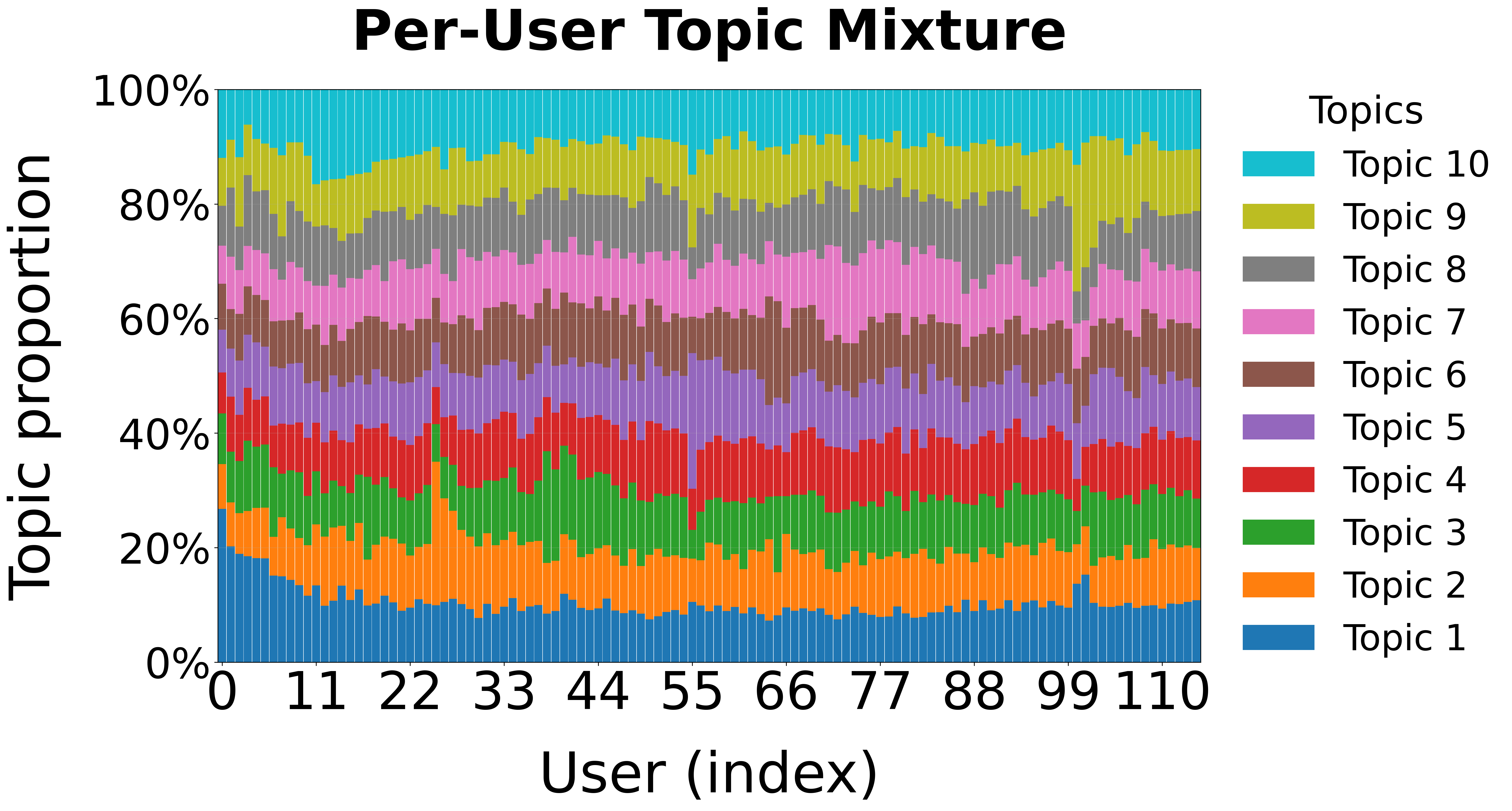}
\end{center}
\caption{Per-user topic mixture distributions for the 115 active Chatbot Arena users (FastTopic). Each bar represents one user, and colors denote latent topic proportions. Topic profiles vary substantially across users, ranging from highly concentrated to broadly distributed.}
\label{fig::fasttopic_results}
\end{figure}

Topic modeling analysis shows substantial heterogeneity in the topics users query LLM models. In \Cref{fig::fasttopic_results} we present the inferred user-level topic proportions for the most active Chatbot Arena users. In addition, 
\Cref{fig::11473GLOCOM} up to \Cref{fig::9965GLOCOM} in Appendix \Cref{appendix:topic_modeling}  include three complementary views: an intertopic distance map showing how topics cluster in 2D space via multidimensional scaling, the top-30 most salient terms for each identified topic, and marginal topic distributions indicating the relative prevalence of each topic in that user's queries.

The visualizations reveal important differences in topical focus across users. For instance, user 1338 shows concentrated interest in a small set of specific topics, with only 4 main topic clusters emerging from their queries. In contrast, user 13046 demonstrates remarkably broad topical coverage, with over 20 diverse topics spanning their query history. This nearly five-fold difference in topical breadth illustrates the spectrum from specialists to generalists in our user population. Beyond breadth, users also differ in the nature of their topical interests: user 1338's queries concentrate heavily on history, government, and civilization-related topics, suggesting domain-specific expertise or interest; meanwhile, user 15085's queries span science, technology, and general knowledge topics more evenly, indicating an exploratory approach.

\begin{figure}[t]
\begin{center}
\includegraphics[width=\columnwidth]{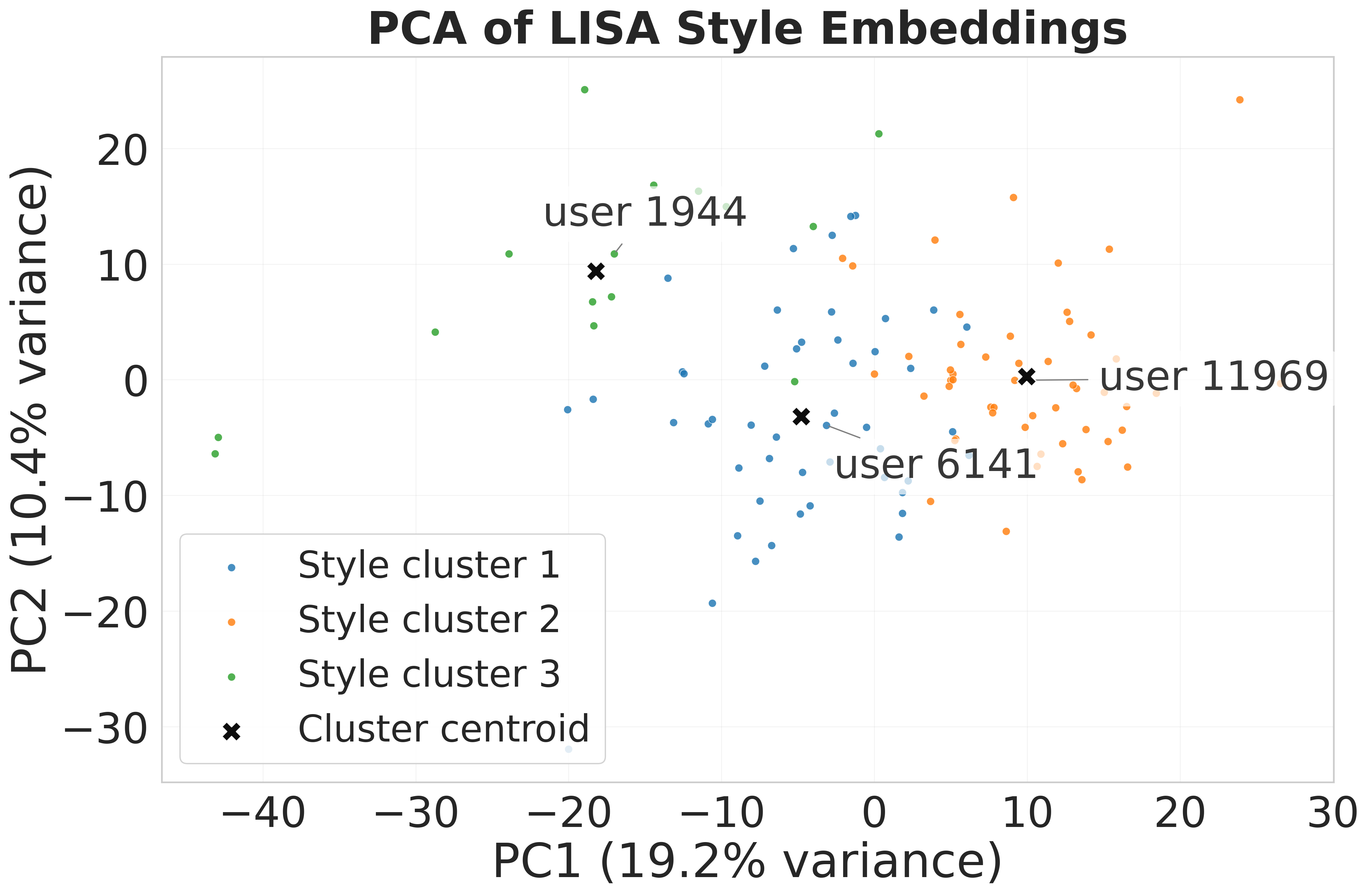}
\end{center}
\caption{Two-dimensional PCA projection of 768-dimensional LISA style feature vectors for the 115 active Chatbot Arena users.}
\label{fig::stylepca}
\end{figure}

We also present the topic coherence-diversity (TC-TD) tradeoff curves for these users in \Cref{fig:topic_coherence_diversity_1} - \Cref{fig:topic_coherence_diversity_3}.  %
The optimal number of topics per user selected to balance high coherence and diversity varies substantially across users (5-25).
This variation quantitatively confirms that users differ not just in what they ask about, but in the fundamental diversity of their information needs.

\subsection{Style Heterogeneity Across Users}

\subsubsection{LISA Style Embeddings}

\begin{figure}[t]
\begin{center}
\includegraphics[width=\columnwidth]{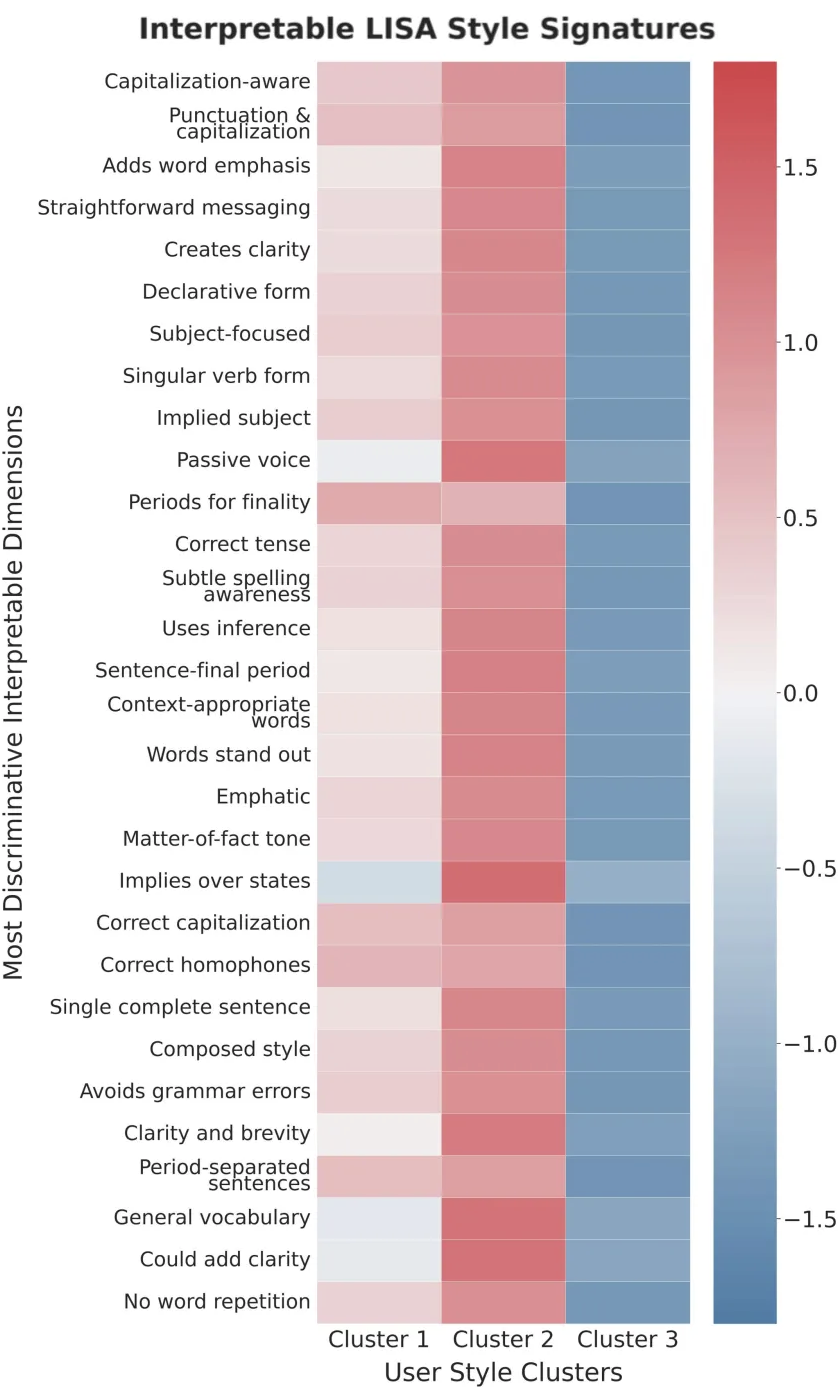}
\end{center}
\caption{Most discriminative LISA style dimensions across the three user style clusters identified via PCA.}
\label{fig::clusterstyleanalysis}
\end{figure}

To further investigate style heterogeneity across users, we apply PCA to the 768-dimensional LISA style feature vectors and present the results in \Cref{fig::stylepca}. The first two principal components explain 19.2\% and 10.4\% of the total variance respectively, and projecting users onto this 2D space reveals three distinct style clusters whose centroids are represented by users 1944, 6141, and 11969. \Cref{fig::clusterstyleanalysis} further characterizes these clusters through their most discriminative LISA dimensions: Cluster 2 is distinguished by strongly positive loadings on dimensions related to structured, formal writing such as correct capitalization, use of declarative forms, and composed style; Cluster 3 shows markedly negative loadings on these same dimensions, suggesting a more fragmented and informal query style, while Cluster 1 occupies an intermediate profile. These results confirm that user writing styles fall into interpretable groupings reflecting systematic differences in how users formulate their queries.

Next we analyze which stylistic dimensions score highest for different users to identify predominant characteristics of their writing style and distinct signatures in how they formulate their queries. In Appendix \ref{appendix:style_analysis} - \Cref{fig::11473LISA} up to \Cref{fig::9965LISA} we present the top 20 LISA style dimensions identified across various users. We observe significant differences in the tone they use and their level of formality. For example, user 11473 predominantly exhibits no neutral tone (scoring 0.90), frequently omits articles (0.83), and uses incomplete sentence structures (0.83). Similarly, user 13046 also shows a not neutral, unenthusiastic tone (0.89 and 0.83 respectively) with incomplete sentences. However, user 6467 takes a notably different approach with blunt, direct writing (0.92) combined with a neutral tone (0.91) and a tendency to avoid numerical references (0.88). These differences suggest varying levels of formality and directness in how users communicate with LLMs.

Sentence structure patterns also distinguish users from one another. For instance, user 1338 tends toward queries with few commas or periods and incomplete grammatical structure, suggesting a telegraphic query style that prioritizes brevity; user 15085 shares the incomplete structure and blunt tone but additionally avoids motion-related words, while user 257 shows a pattern of omitting articles and avoiding motion perception vocabulary. 

These structural differences reflect different cognitive approaches to query formulation: some users craft complete sentences while others use keyword-style queries. These patterns suggest that stylistic choices in query formulation are systematic rather than random, and reflect users' professional backgrounds, communication preferences and interaction strategies with AI systems.

\begin{figure}[t]
\begin{center}
\includegraphics[width=\columnwidth]{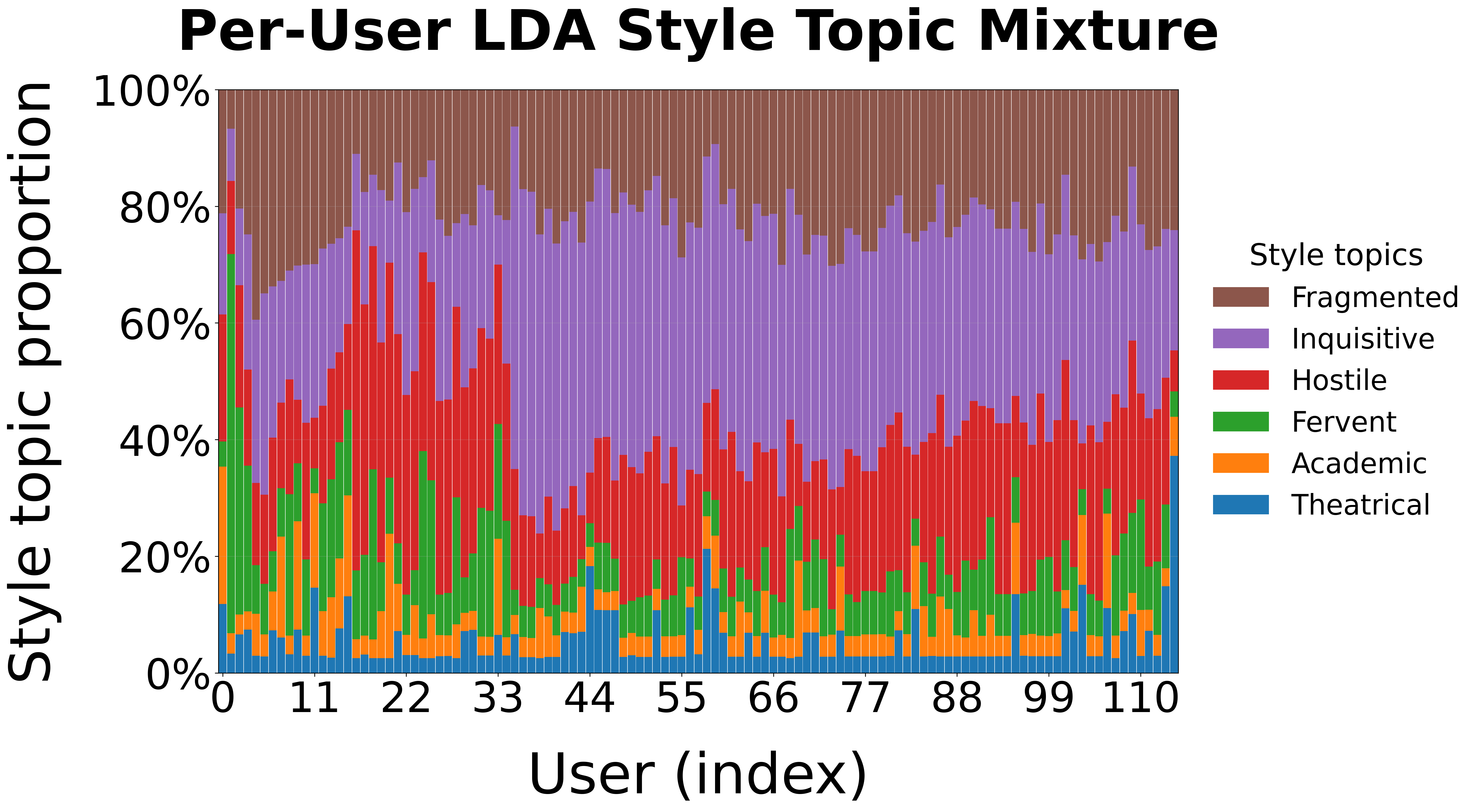}
\end{center}
\caption{Proportion of LDA-derived style topics per user, illustrating the heterogeneity of writing style mixtures across Chatbot Arena active users. }
\label{fig::ldastyletopics}
\end{figure}

\subsubsection{Inferring Stylistic Topics}

To provide a higher-level organization of these stylistic patterns, we apply LDA to the LISA style dimensions. We identify 6 meta-styles that capture common patterns across users, and span a broad spectrum of communication approaches. We present results in \Cref{fig::ldastyletopics} and \Cref{table::StyleLabel}.

The \textit{\textbf{Theatrical}} style combines dramatic flair with confident, evocative writing that employs figurative language strategically. The \textit{\textbf{Academic}} style emphasizes formal scholarly conventions, precision, and complex sentence structures with specialized vocabulary. The \textit{\textbf{Fervent}} style is emotionally charged and intense, using vivid description and passionate language. In contrast, the \textit{\textbf{Hostile}} style takes an aggressively confrontational approach characterized by bluntness, rudeness, and dismissiveness. The \textit{\textbf{Inquisitive}} style reflects a professional, exploratory approach focused on questioning with formal precision. Finally, the \textit{\textbf{Fragmented}} style features incomplete sentences, sparse punctuation, and an informal, unpolished quality. This variation demonstrates that style topics capture meaningful differences in how users formulate queries, and  provide a compact representation of stylistic heterogeneity across the user population.

\subsubsection{HypoGeniC Style Characterizations}

In addition to  quantitative measures, we also present qualitative characterizations of users writing style inferred using HypoGeniC. In Appendix \Cref{appendix:style_analysis}, \Cref{table::user_9965} - \Cref{table::arena_user_5203} show the hypotheses generated by HypoGeniC, with 10 hypotheses per user describing their writing style.

For example, user 9965 is characterized as comfortable with ambiguity and uncertainty, providing lengthy detailed responses with tangential thoughts, creative and humorous language employing wordplay, and a fondness for metaphors and figurative language. In contrast, user 257 is characterized as comfortable with mathematical and logical problems, showing strong interest in history and current events, with a casual, informal tone using colloquial language. User 15085 exhibits playful, whimsical language with enjoyment of wordplay, delivered through a casual conversational tone featuring humor and irony, with a focus on brevity and concision contrasting with user 9965's verbose style. 

Across all users, the HypoGeniC hypotheses identify meaningful dimensions of variation including formality versus casualness, creativity versus precision, humor versus seriousness, and verbosity versus conciseness. These qualitative characterizations complement the quantitative LISA embeddings by providing interpretable descriptions of users' writing style.

\subsection{Predicting User-Specific Model Rankings}

The substantial topic and style heterogeneity documented above raises a natural question: can these user-level features predict individual model preferences? We address this next.

\begin{table}[t]
\begin{center}
\scalebox{0.9}{
  \begin{tabular}{ l c c}
  \toprule
    \textbf{Model} & \textbf{Mean-Predictor} & \textbf{Topic + Style} \\ 
    & \textbf{Baseline} & \textbf{(Ours)} \\
    \midrule
    ELO & 0.688 & 0.450 ($35\% \downarrow$) \\ 
    Bradley-Terry & 0.510 & 0.450 ($12\% \downarrow$)\\
    \bottomrule
  \end{tabular}}
\end{center}
\caption{MAE on the validation set for the mean-predictor baseline versus the combined topic and style regression model, under both ELO and Bradley-Terry target formulations. Lower is better.} %
\label{table::regression}
\end{table}

Using the combined topic and style representation for each user, the regression models achieves similar predictive performance across both target formulations: the best ELO and BT regression models reach a scaled MAE of $0.450$ on the validation set; see \Cref{table::regression}. 
Compared to a mean-predictor baseline which predicts every user's ranking as the average ranking vector across training users, essentially mirroring what an aggregate benchmark would prescribe, our model reduces scaled MAE from 0.688 to 0.450 for ELO ($35\%$ improvement)  and from 0.510 to 0.450 for BT ($12\%$ improvement). These results show that topic and style features carry genuine personalization signal beyond what population-level averages can provide.

\section{Discussion}
\label{sec::discussion}

\paragraph{Dimensions of User Query Heterogeneity}
The dimensions along which users diverge (topical specialization, communication style, and query complexity) map onto distinct model capability profiles. A specialist querying technical domains in formal language likely benefits from a model optimized for depth and precision, while a generalist using fragmented, playful queries may be better served by a model with strong conversational flexibility. This suggests that personalized benchmarking is not only about measuring preference divergence, but about understanding which model capabilities are actually being selected for by different user populations -- a question aggregate benchmarks cannot answer by design.

\paragraph{Limitations of Aggregate Benchmarks} The near-complete absence of alignment between individual and aggregate Bradley-Terry rankings for the majority of users means that a single global leaderboard is not merely imprecise, \textit{it is actively misleading for model selection}. The substantial heterogeneity we document has direct implications for how LLM evaluation is conducted.
A highly-ranked model might achieve its position through moderate performance across all user types rather than excellence for any particular group. Models optimized for aggregate performance may not serve individual segments well, and development resources focused on aggregate metrics may miss opportunities to better serve specific populations. Furthermore, because ELO's incremental update mechanism averages preference signals across comparisons, its moderate population-level correlation likely reflects smoothing toward the aggregate rather than genuine individual alignment; the true degree of preference divergence across users may be even larger than ELO-based estimates suggest.

\paragraph{Towards Practical Personalized Benchmarking}
Translating personalized benchmarking into practice need not require fundamental changes to existing infrastructure. A user's topic and style profile can be inferred from a small number of queries and used to match them to appropriate models based on the preferences of similar users, without extensive preference elicitation. Rather than a single global leaderboard, benchmarks could report separate model rankings for different types of users; for example, one ranking for users who ask technical questions in formal language, and another for users who explore creative topics more casually, helping developers identify which model best serves their intended audience. These approaches can be layered on top of existing platforms like Chatbot Arena by enriching interaction logs with topic and style annotations. By implementing these changes, personalized benchmarking becomes a practically achievable near-term goal.

\section{Conclusion}
\label{sec::conclusion}

We have introduced personalized benchmarking for LLMs through comprehensive analysis of user query patterns and model ranking correlations. Our analysis of $115$ active Chatbot Arena users demonstrates that users exhibit substantial heterogeneity in the topics they query, their writing styles, and critically, in their model preferences as measured by personalized rankings.

We identify interpretable dimensions along which users differ, from topical specialization ($4-20+$ topics) to communication style (formal to casual, creative to precise). Our ranking correlation analysis provides  quantitative evidence that these differences translate to divergent model preferences: while ELO-based individual rankings show moderate correlation with global rankings ($\rho = 0.43$), Bradley-Terry rankings reveal dramatic divergence ($\rho = 0.04$), with $57\%$ of users showing near zero or negative correlation with the aggregate -- meaning \textit{aggregate rankings fail to predict individual preferences for the majority of users}. Furthermore, we demonstrate that a compact combination of topic and style features provides a useful feature space for predicting user-specific model rankings, suggesting a lightweight path toward operationalizing personalized benchmarking in practice.

Our findings challenge the one-size-fits-all paradigm of current LLM benchmarking. 
Aggregate benchmarks provide valuable signals about general model capabilities, but they obscure substantial individual heterogeneity that affects real-world user satisfaction. 
Rather than seeking a single ``best'' model, we should acknowledge the diversity of user needs and develop systems that match users to models suited to their preferences and use cases. Existing platforms like Chatbot Arena already collect the interaction data needed to support this paradigm; the key missing ingredient is enrichment with topic and style annotations as developed in this work, making personalized benchmarking a practically achievable near-term goal. As LLMs become increasingly central to how people work and create, ensuring that these systems serve individual needs effectively becomes paramount.

\textbf{Limitations} Our analysis is limited to $115$ Chatbot Arena users with $\ge25$ preference votes, and may miss subtle nuances in tone, intent, or context; findings are correlational rather than causal. Analysis is restricted to English queries, leaving open how user heterogeneity manifests across languages.

\section*{Acknowledgments}
We thank the anonymous reviewers for their insightful comments. This work is
supported in part by NSF grants IIS-2126602, 2302785, 
an award from the Sloan Foundation, and AI research initiatives at the University of Chicago.

\bibliography{main}

\begin{thebibliography}{21}
\providecommand{\natexlab}[1]{#1}

\bibitem[{{Anthropic}(2025)}]{anthropic2025claude45sonnet}
{Anthropic}. 2025.
\newblock Claude 4.5 sonnet.
\newblock \url{https://www.anthropic.com/news/claude-sonnet-4-5}.

\bibitem[{Bai et~al.(2024)Bai, Liu, Bu, He, Liu, Zhou, Lin, Su, Ge, Zheng et~al.}]{bai2024mt}
Ge~Bai, Jie Liu, Xingyuan Bu, Yancheng He, Jiaheng Liu, Zhanhui Zhou, Zhuoran Lin, Wenbo Su, Tiezheng Ge, Bo~Zheng, and 1 others. 2024.
\newblock Mt-bench-101: A fine-grained benchmark for evaluating large language models in multi-turn dialogues.
\newblock In \emph{Proceedings of the 62nd Annual Meeting of the Association for Computational Linguistics (Volume 1: Long Papers)}, pages 7421--7454.

\bibitem[{Bradley and Terry(1952)}]{bradley1952rank}
Ralph~Allan Bradley and Milton~E Terry. 1952.
\newblock Rank analysis of incomplete block designs: I. the method of paired comparisons.
\newblock \emph{Biometrika}, 39(3/4):324--345.

\bibitem[{Chiang et~al.(2024)Chiang, Zheng, Sheng, Angelopoulos, Li, Li, Zhu, Zhang, Jordan, Gonzalez et~al.}]{chiang2024chatbot}
Wei-Lin Chiang, Lianmin Zheng, Ying Sheng, Anastasios~N Angelopoulos, Tianle Li, Dacheng Li, Banghua Zhu, Hao Zhang, Michael~I Jordan, Joseph~E Gonzalez, and 1 others. 2024.
\newblock Chatbot arena: an open platform for evaluating llms by human preference.
\newblock In \emph{Proceedings of the 41st International Conference on Machine Learning}, pages 8359--8388.

\bibitem[{Ehsan et~al.(2021)Ehsan, Liao, Muller, Riedl, and Weisz}]{ehsan2021expanding}
Upol Ehsan, Q~Vera Liao, Michael Muller, Mark~O Riedl, and Justin~D Weisz. 2021.
\newblock Expanding explainability: Towards social transparency in ai systems.
\newblock In \emph{Proceedings of the 2021 CHI conference on human factors in computing systems}, pages 1--19.

\bibitem[{Elo(1967)}]{elo1967proposed}
Arpad~E Elo. 1967.
\newblock The proposed uscf rating system, its development, theory, and applications.
\newblock \emph{Chess life}, 22(8):242--247.

\bibitem[{Garbacea and Tan(2025)}]{garbacea2025hyperalign}
Cristina Garbacea and Chenhao Tan. 2025.
\newblock Hyperalign: Interpretable personalized llm alignment via hypothesis generation.
\newblock \emph{arXiv preprint arXiv:2505.00038}.

\bibitem[{Grattafiori et~al.(2024)Grattafiori, Dubey, Jauhri, Pandey, Kadian, Al-Dahle, Letman, Mathur, Schelten, Vaughan et~al.}]{grattafiori2024llama}
Aaron Grattafiori, Abhimanyu Dubey, Abhinav Jauhri, Abhinav Pandey, Abhishek Kadian, Ahmad Al-Dahle, Aiesha Letman, Akhil Mathur, Alan Schelten, Alex Vaughan, and 1 others. 2024.
\newblock The llama 3 herd of models.
\newblock \emph{arXiv preprint arXiv:2407.21783}.

\bibitem[{Lambert(2025)}]{lambert2025reinforcement}
Nathan Lambert. 2025.
\newblock Reinforcement learning from human feedback.
\newblock \emph{arXiv preprint arXiv:2504.12501}.

\bibitem[{Li et~al.(2025)Li, Chiang, Frick, Dunlap, Wu, Zhu, Gonzalez, and Stoica}]{li2025crowdsourced}
Tianle Li, Wei-Lin Chiang, Evan Frick, Lisa Dunlap, Tianhao Wu, Banghua Zhu, Joseph~E Gonzalez, and Ion Stoica. 2025.
\newblock From crowdsourced data to high-quality benchmarks: Arena-hard and benchbuilder pipeline.
\newblock In \emph{International Conference on Machine Learning}, pages 34209--34231. PMLR.

\bibitem[{Li et~al.(2023)Li, Zhang, Dubois, Taori, Gulrajani, Guestrin, Liang, and Hashimoto}]{alpaca_eval}
Xuechen Li, Tianyi Zhang, Yann Dubois, Rohan Taori, Ishaan Gulrajani, Carlos Guestrin, Percy Liang, and Tatsunori~B. Hashimoto. 2023.
\newblock Alpacaeval: An automatic evaluator of instruction-following models.
\newblock \url{https://github.com/tatsu-lab/alpaca_eval}.

\bibitem[{Malik et~al.(2024)Malik, Jiang, and Chai}]{malik2024empirical}
Manuj Malik, Jing Jiang, and Kian Ming~A Chai. 2024.
\newblock An empirical analysis of the writing styles of persona-assigned llms.
\newblock In \emph{Proceedings of the 2024 conference on empirical methods in natural language processing}, pages 19369--19388.

\bibitem[{OpenAI(2025)}]{openai2025gpt5}
OpenAI. 2025.
\newblock Chatgpt (gpt-5).
\newblock \url{https://openai.com/index/introducing-gpt-5/}.

\bibitem[{Ouyang et~al.(2022)Ouyang, Wu, Jiang, Almeida, Wainwright, Mishkin, Zhang, Agarwal, Slama, Ray et~al.}]{ouyang2022training}
Long Ouyang, Jeffrey Wu, Xu~Jiang, Diogo Almeida, Carroll Wainwright, Pamela Mishkin, Chong Zhang, Sandhini Agarwal, Katarina Slama, Alex Ray, and 1 others. 2022.
\newblock Training language models to follow instructions with human feedback.
\newblock \emph{Advances in neural information processing systems}, 35:27730--27744.

\bibitem[{Patel et~al.(2023)Patel, Rao, Kothary, McKeown, and Callison-Burch}]{patel2023learning}
Ajay Patel, Delip Rao, Ansh Kothary, Kathleen McKeown, and Chris Callison-Burch. 2023.
\newblock Learning interpretable style embeddings via prompting llms.
\newblock In \emph{Findings of the Association for Computational Linguistics: EMNLP 2023}, pages 15270--15290.

\bibitem[{Reimers and Gurevych(2019)}]{reimers2019sentence}
Nils Reimers and Iryna Gurevych. 2019.
\newblock Sentence-bert: Sentence embeddings using siamese bert-networks.
\newblock In \emph{Proceedings of the 2019 conference on empirical methods in natural language processing and the 9th international joint conference on natural language processing (EMNLP-IJCNLP)}, pages 3982--3992.

\bibitem[{Rousseeuw(1987)}]{rousseeuw1987silhouettes}
Peter~J Rousseeuw. 1987.
\newblock Silhouettes: a graphical aid to the interpretation and validation of cluster analysis.
\newblock \emph{Journal of computational and applied mathematics}, 20:53--65.

\bibitem[{Spearman(1904)}]{spearman1904proof}
Charles Spearman. 1904.
\newblock The proof and measurement of association between two things.
\newblock \emph{The American Journal of Psychology}, 15(1):72--101.

\bibitem[{Wu et~al.(2024)Wu, Nguyen, Zhang, Wang, and Luu}]{wu2024fastopic}
Xiaobao Wu, Thong Nguyen, Delvin~C Zhang, William~Y Wang, and Anh~T Luu. 2024.
\newblock Fastopic: Pretrained transformer is a fast, adaptive, stable, and transferable topic model.
\newblock \emph{Advances in Neural Information Processing Systems}, 37:84447--84481.

\bibitem[{Zheng et~al.(2023)Zheng, Chiang, Sheng, Zhuang, Wu, Zhuang, Lin, Li, Li, Xing et~al.}]{zheng2023judging}
Lianmin Zheng, Wei-Lin Chiang, Ying Sheng, Siyuan Zhuang, Zhanghao Wu, Yonghao Zhuang, Zi~Lin, Zhuohan Li, Dacheng Li, Eric Xing, and 1 others. 2023.
\newblock Judging llm-as-a-judge with mt-bench and chatbot arena.
\newblock \emph{Advances in neural information processing systems}, 36:46595--46623.

\bibitem[{Zhou et~al.(2024)Zhou, Liu, Srivastava, Mei, and Tan}]{zhou2024hypothesis}
Yangqiaoyu Zhou, Haokun Liu, Tejes Srivastava, Hongyuan Mei, and Chenhao Tan. 2024.
\newblock Hypothesis generation with large language models.
\newblock In \emph{Proceedings of the 1st Workshop on NLP for Science (NLP4Science)}, pages 117--139.

\end{thebibliography}

\appendix

\section{Appendix}

\subsection{Topic Modeling Results}
\label{appendix:topic_modeling}

\begin{figure*}[t]
\begin{center}
\includegraphics[width=\linewidth]{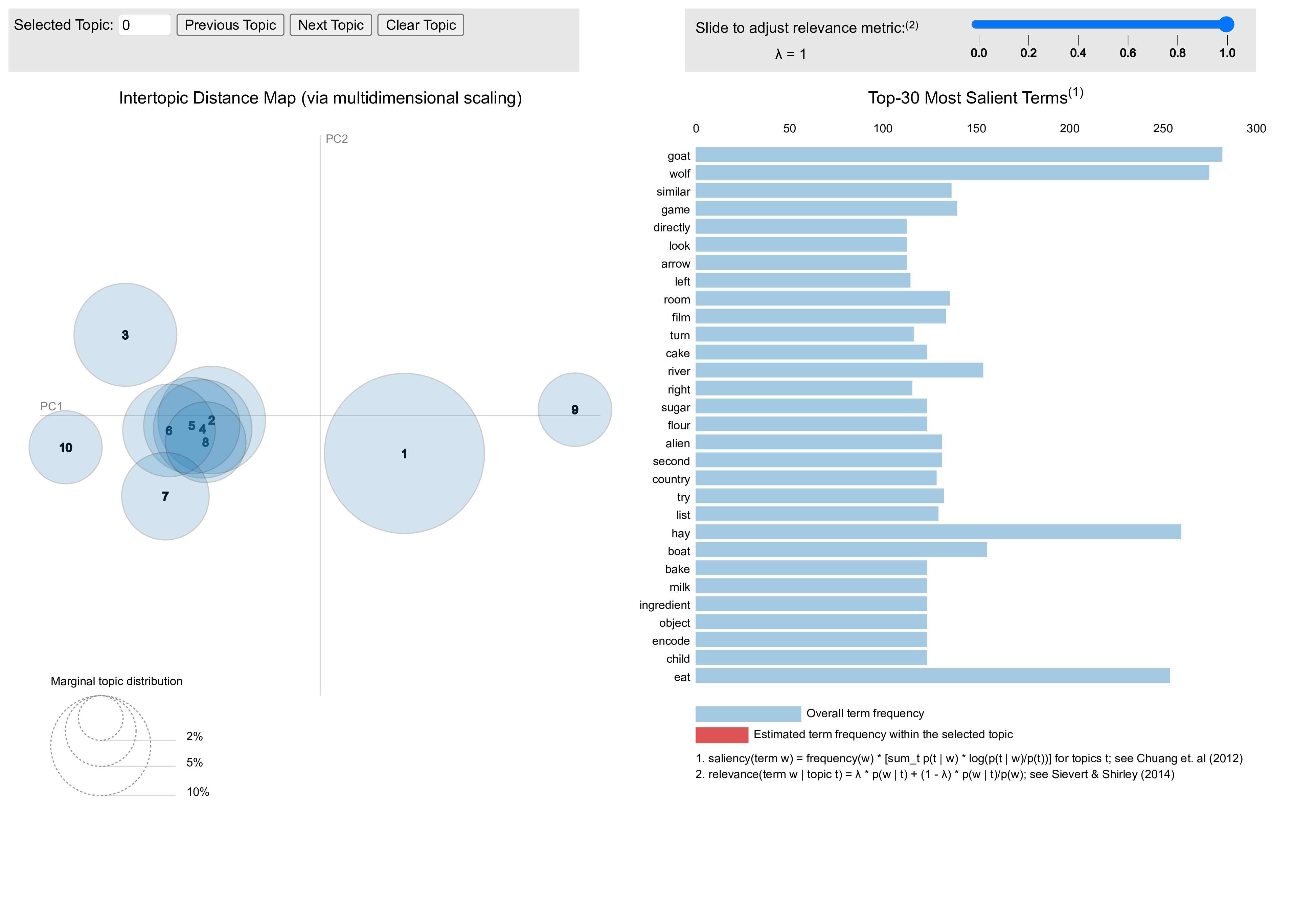}
\end{center}
\caption{Topic analysis results for Chatbot Arena user 11473}
\label{fig::11473GLOCOM}
\end{figure*}

\begin{figure*}[t]
\begin{center}
\includegraphics[width=\linewidth]{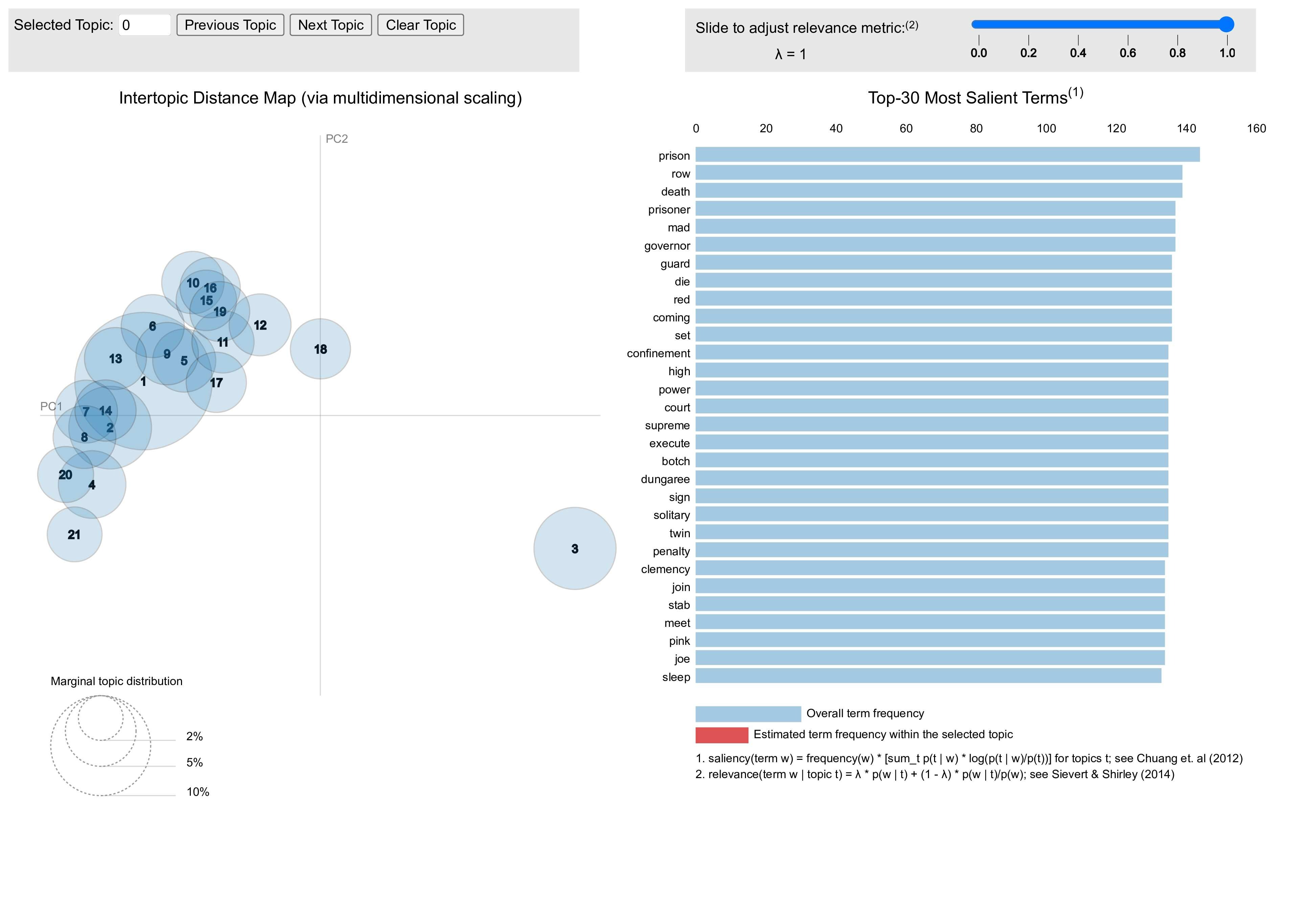}
\end{center}
\caption{Topic analysis results for Chatbot Arena user 13046}
\label{fig::13046GLOCOM}
\end{figure*}

\begin{figure*}[t]
\begin{center}
\includegraphics[width=\linewidth]{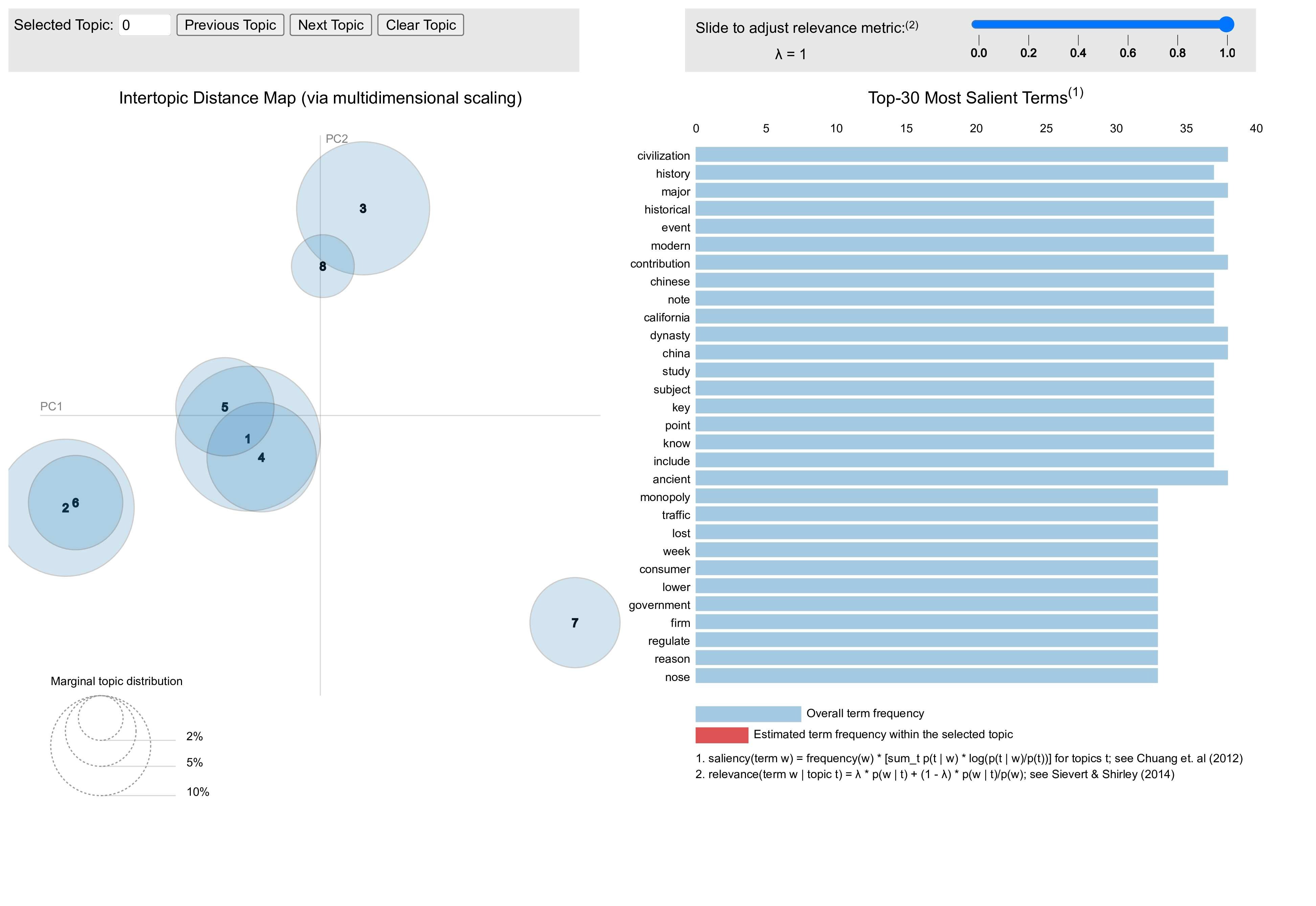}
\end{center}
\caption{Topic analysis results for Chatbot Arena user 1338}
\label{fig::1338GLOCOM}
\end{figure*}

\begin{figure*}[t]
\begin{center}
\includegraphics[width=\linewidth]{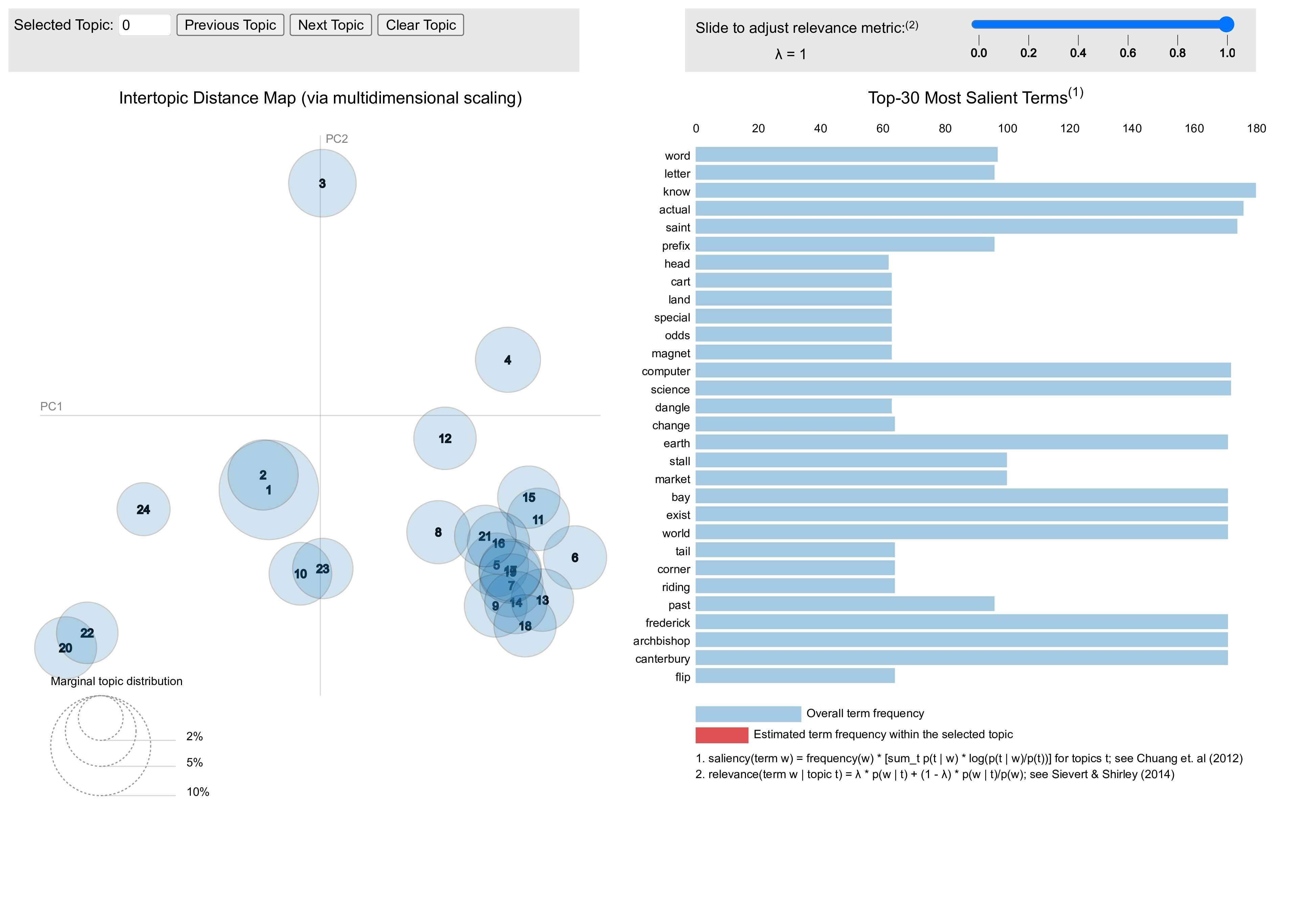}
\end{center}
\caption{Topic analysis results for Chatbot Arena user 15085}
\label{fig::15085GLOCOM}
\end{figure*}

\begin{figure*}[t]
\begin{center}
\includegraphics[width=\linewidth]{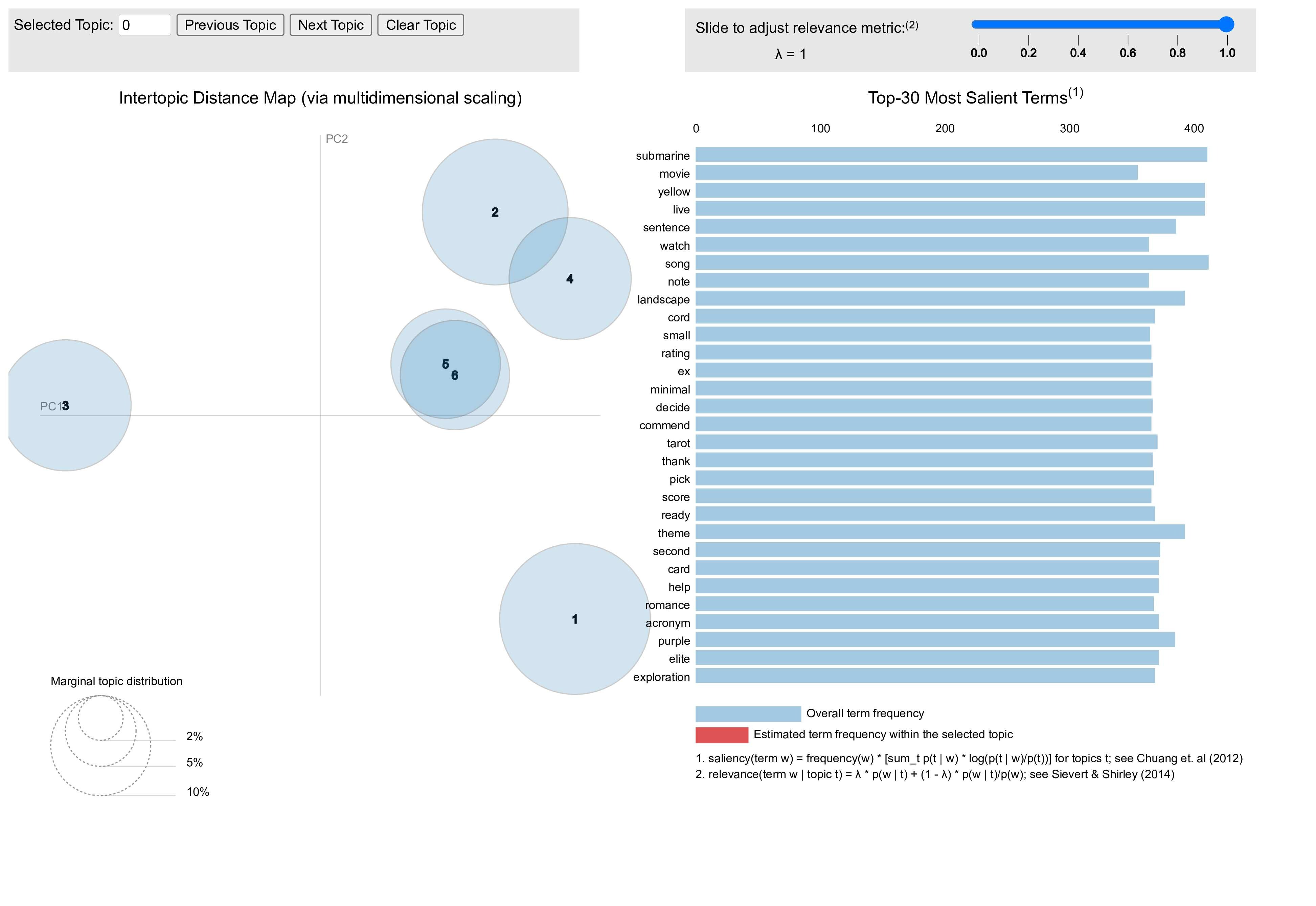}
\end{center}
\caption{Topic analysis results for Chatbot Arena user 257}
\label{fig::257GLOCOM}
\end{figure*}

\begin{figure*}[t]
\begin{center}
\includegraphics[width=\linewidth]{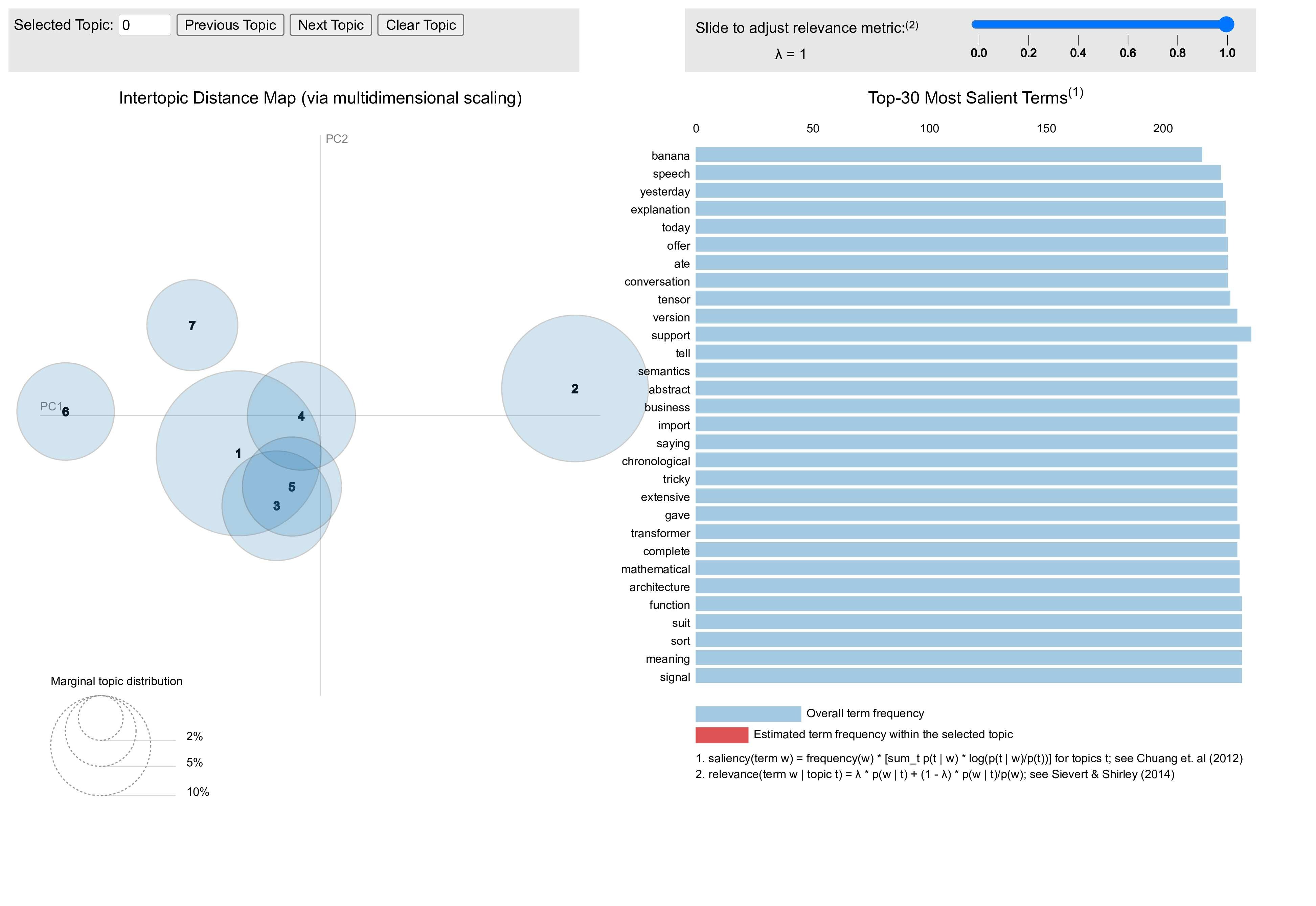}
\end{center}
\caption{Topic analysis results for Chatbot Arena user 3820}
\label{fig::3820GLOCOM}
\end{figure*}

\begin{figure*}[t]
\begin{center}
\includegraphics[width=\linewidth]{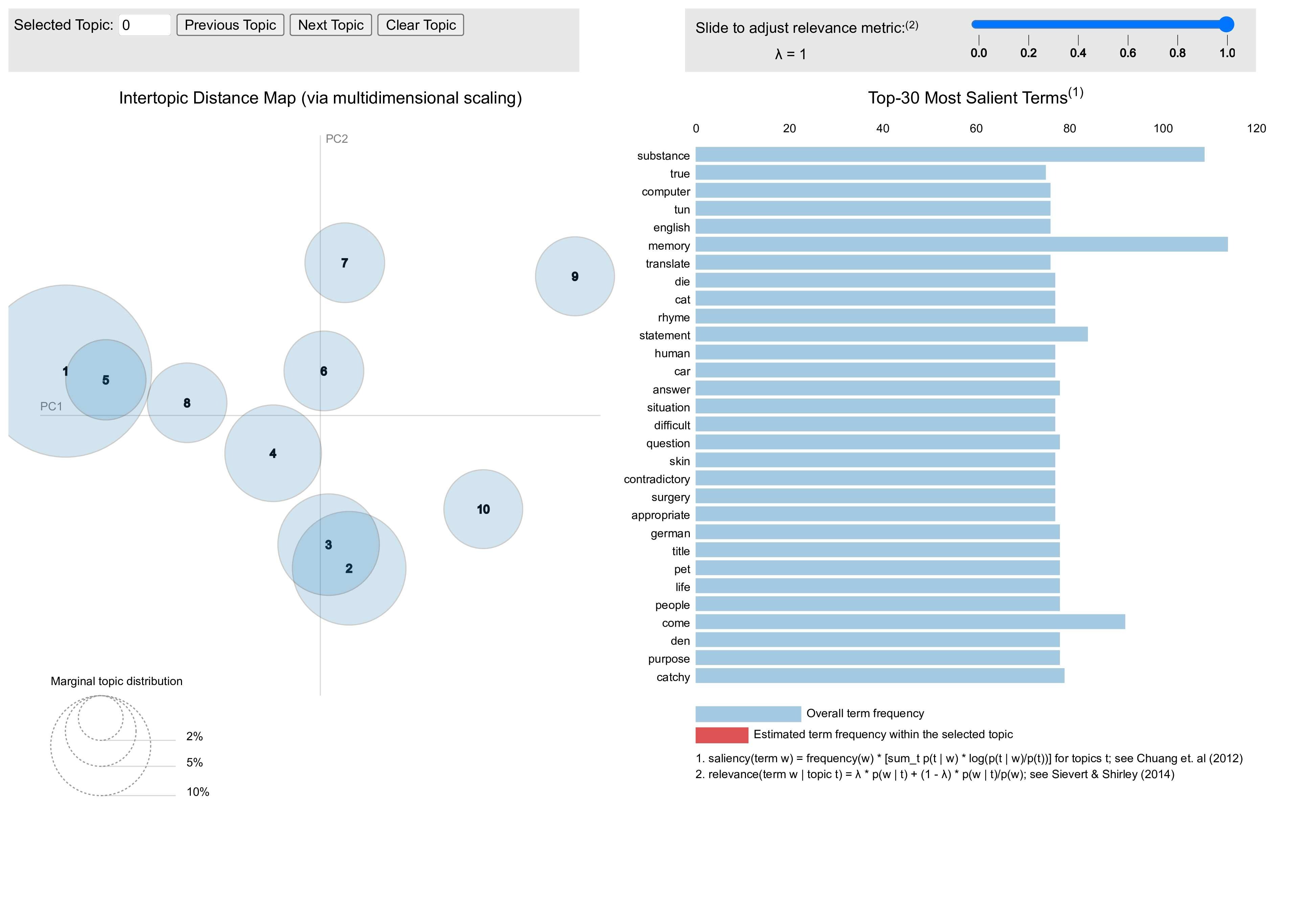}
\end{center}
\caption{Topic analysis results for Chatbot Arena user 6467}
\label{fig::6467GLOCOM}
\end{figure*}

\begin{figure*}[t]
\begin{center}
\includegraphics[width=\linewidth]{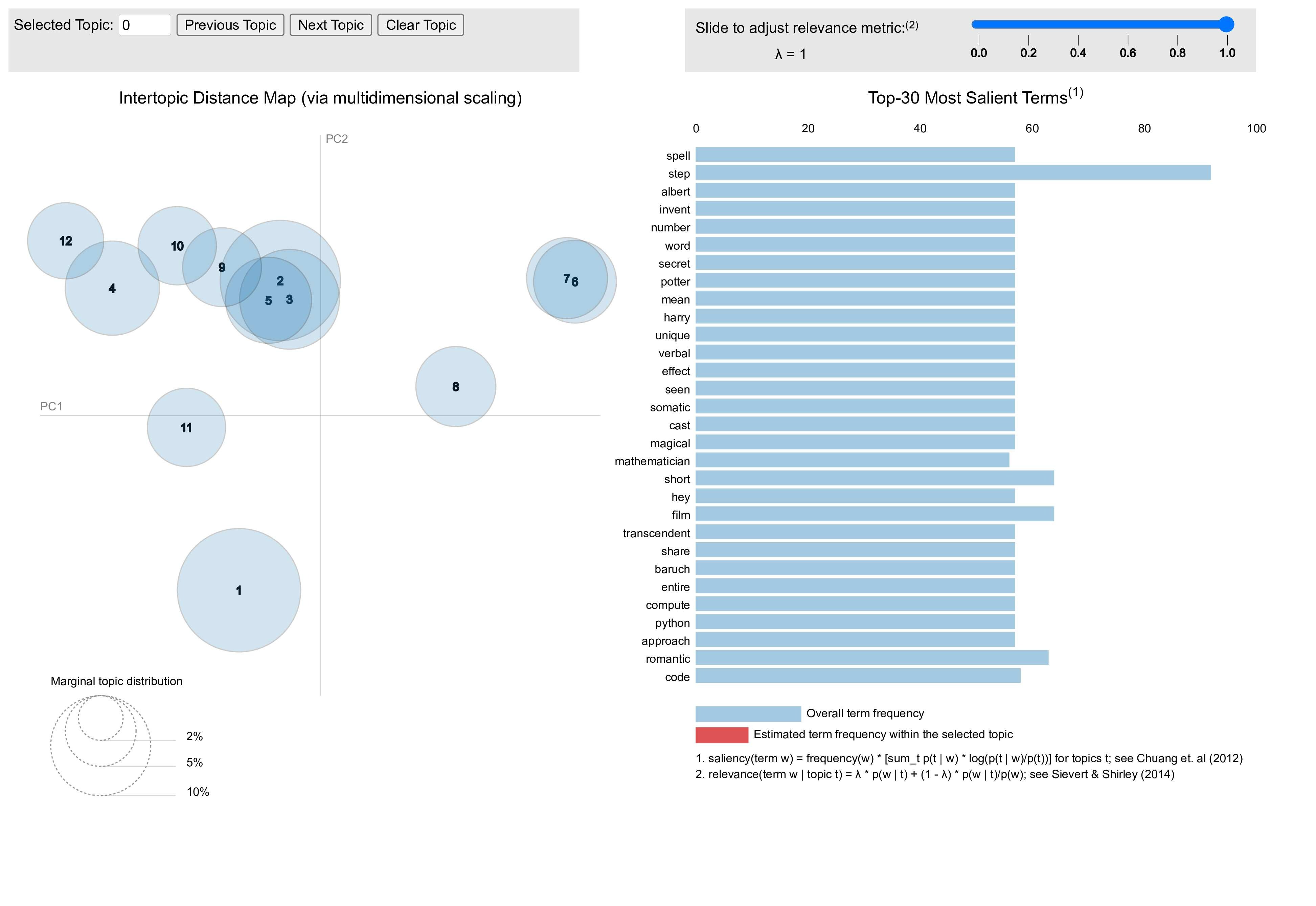}
\end{center}
\caption{Topic analysis results for Chatbot Arena user 6568}
\label{fig::6568GLOCOM}
\end{figure*}

\begin{figure*}[t]
\begin{center}
\includegraphics[width=\linewidth]{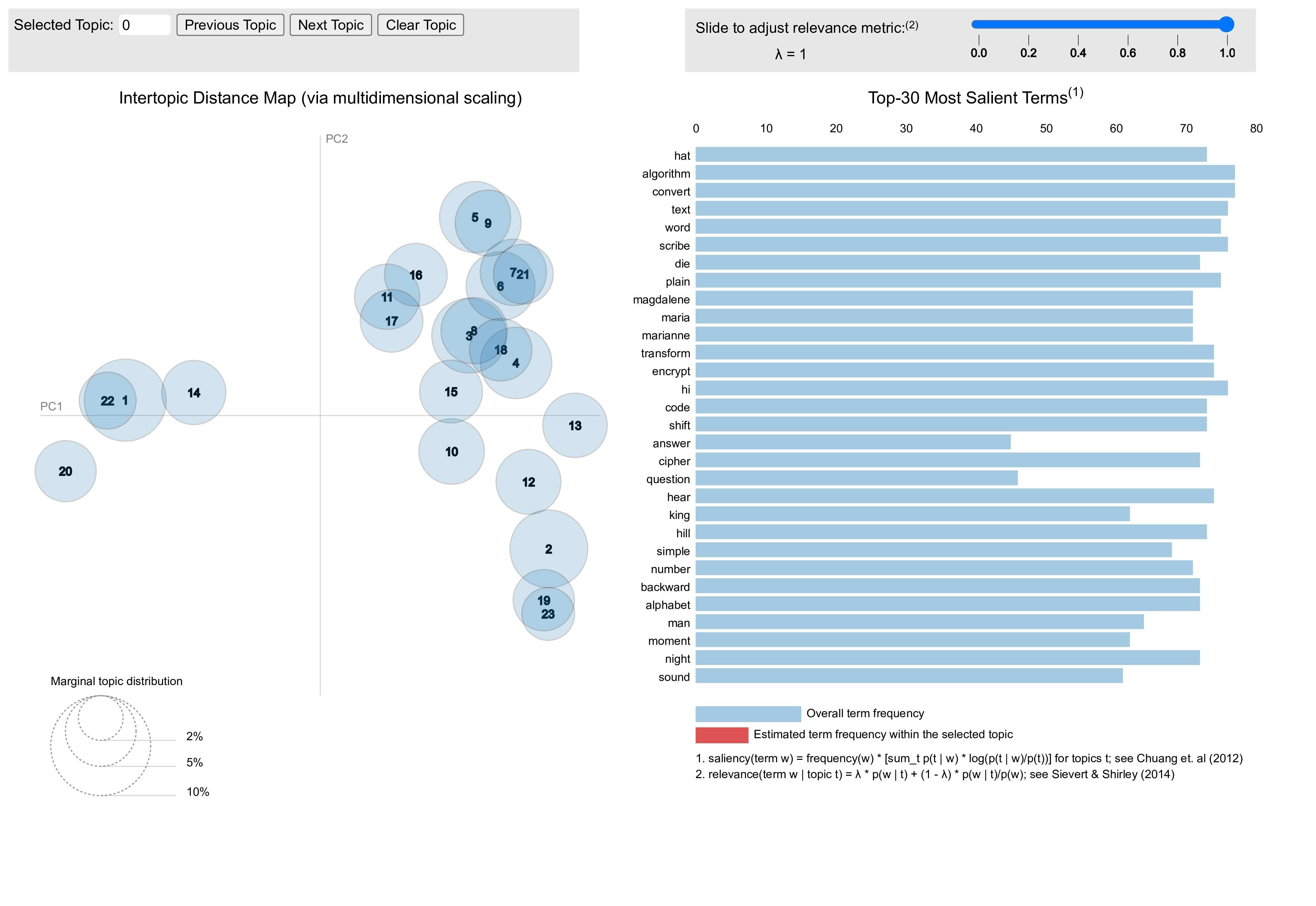}
\end{center}
\caption{Topic analysis results for Chatbot Arena user 9676}
\label{fig::9676GLOCOM}
\end{figure*}

\begin{figure*}[t]
\begin{center}
\includegraphics[width=\linewidth]{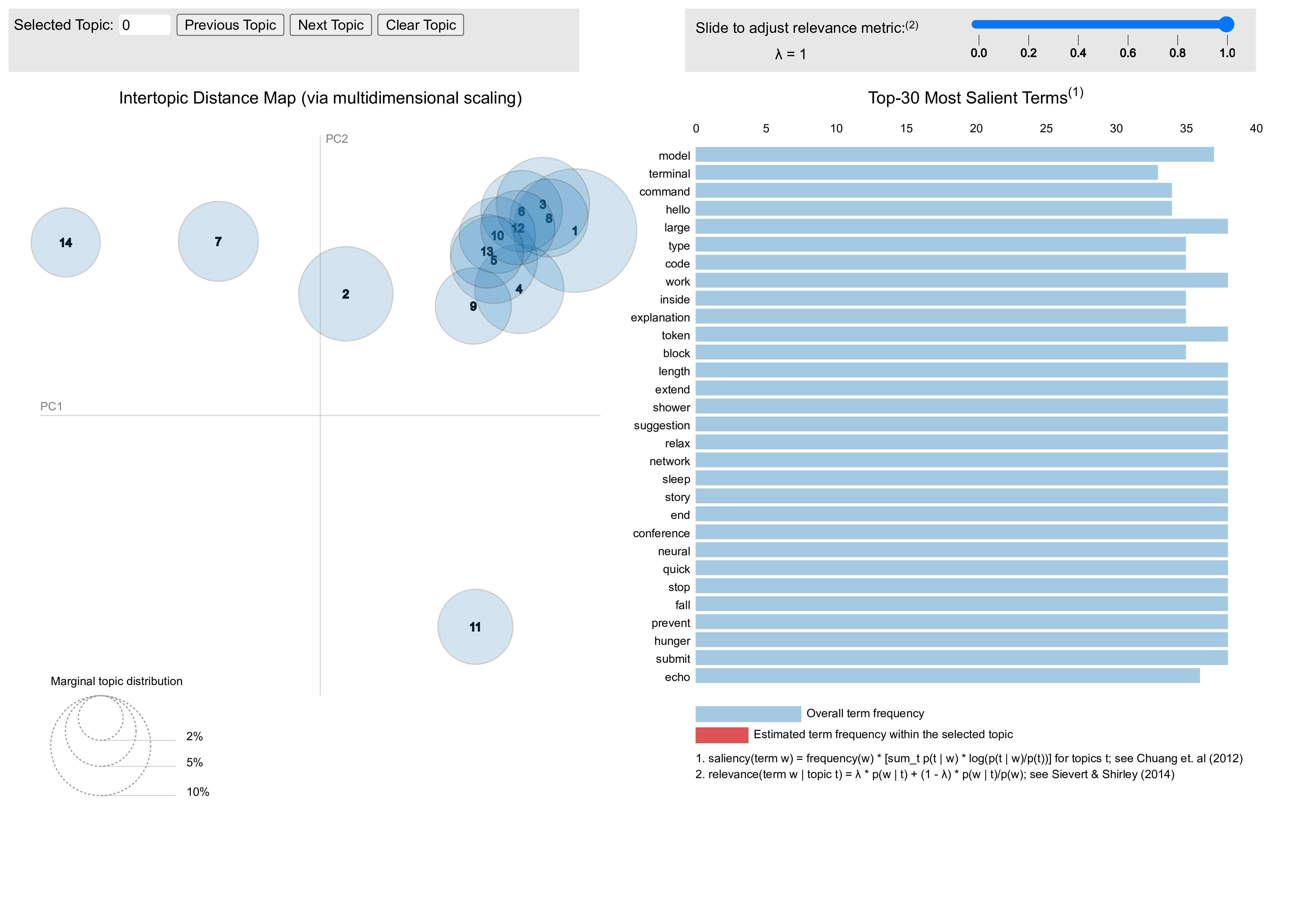}
\end{center}
\caption{Topic analysis results for Chatbot Arena user 973}
\label{fig::973GLOCOM}
\end{figure*}

\begin{figure*}[t]
\begin{center}
\includegraphics[width=\linewidth]{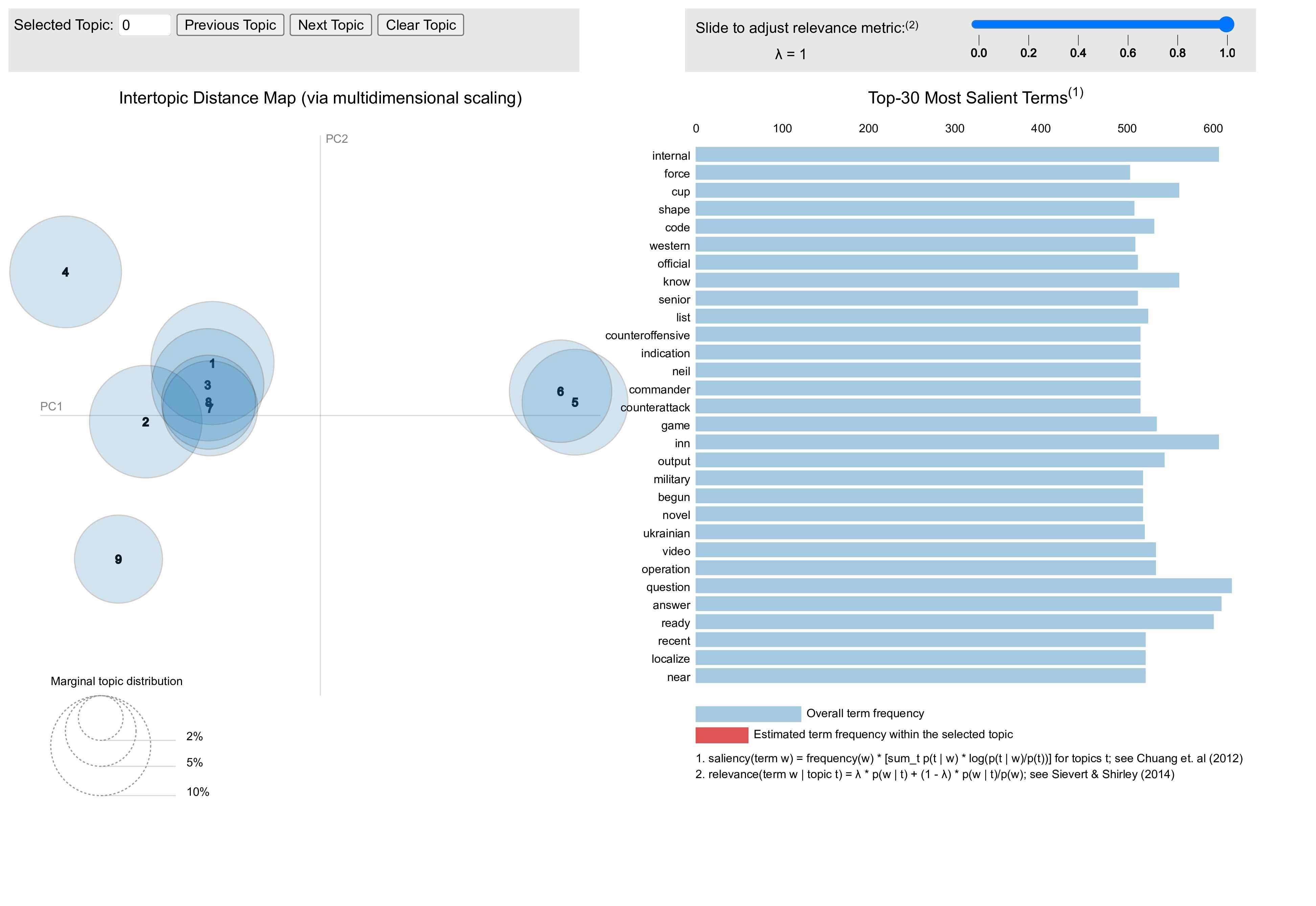}
\end{center}
\caption{Topic analysis results for Chatbot Arena user 9965}
\label{fig::9965GLOCOM}
\end{figure*}

\begin{figure}[htbp]
     \centering
     \begin{subfigure}[b]{0.45\textwidth}
         \centering
         \includegraphics[width=\textwidth]{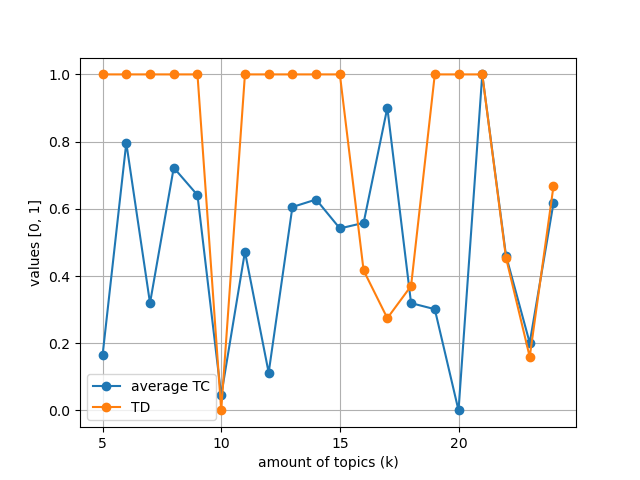}
         \caption{Chatbot Arena user 13046}
     \end{subfigure}
     \hfill
     \begin{subfigure}[b]{0.45\textwidth}
         \centering
         \includegraphics[width=\textwidth]{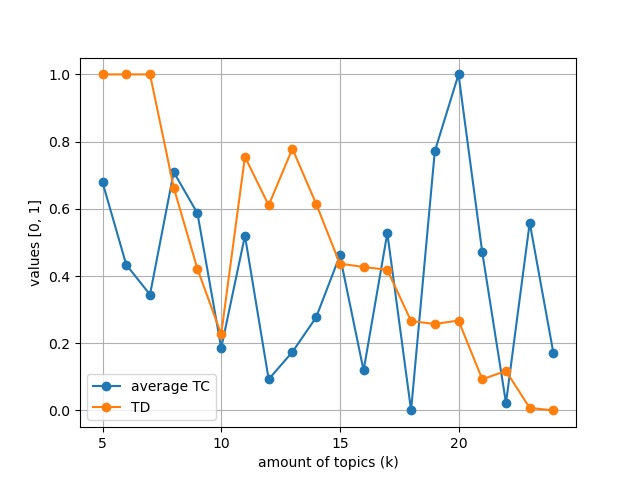}
         \caption{Chatbot Arena user 1338}
     \end{subfigure}

     \vspace{1em} %

     \begin{subfigure}[b]{0.45\textwidth}
         \centering
         \includegraphics[width=\textwidth]{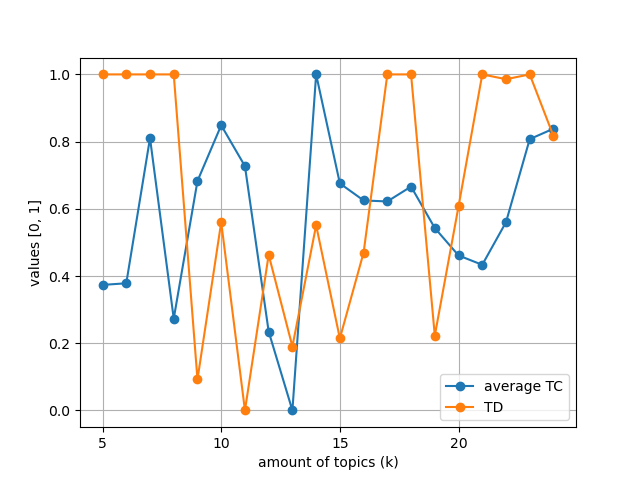}
         \caption{Chatbot Arena user 15085}
     \end{subfigure}
     \hfill
     \begin{subfigure}[b]{0.45\textwidth}
         \centering
         \includegraphics[width=\textwidth]{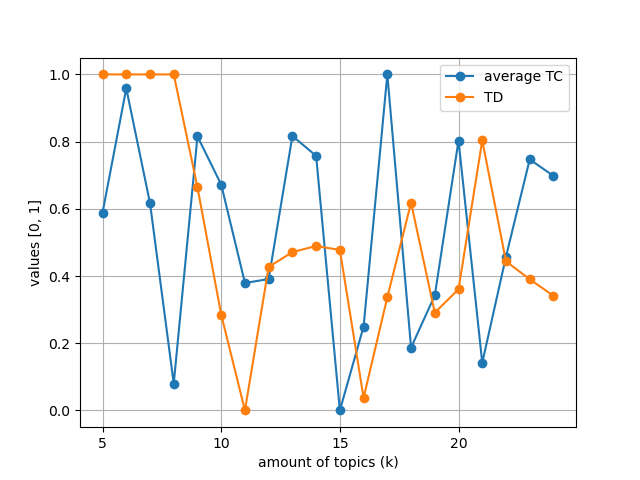}
         \caption{Chatbot Arena user 257}
     \end{subfigure}
     
     \caption{Topic Coherence-Diversity tradeoff curves.}
     \label{fig:topic_coherence_diversity_1}
\end{figure}

\begin{figure}[htbp]
     \centering
     \begin{subfigure}[b]{0.45\textwidth}
         \centering
         \includegraphics[width=\textwidth]{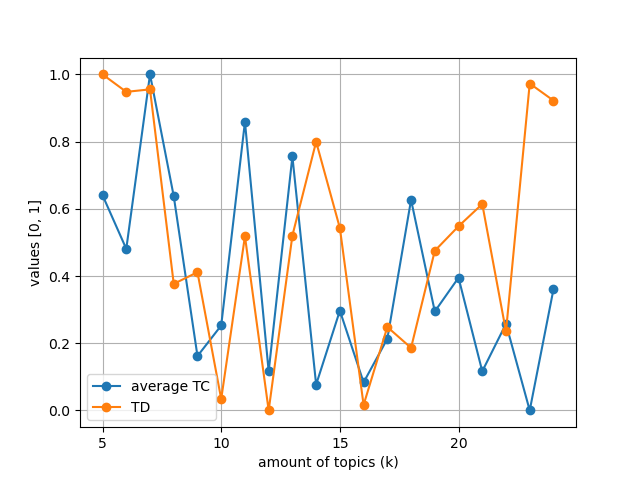}
         \caption{Chatbot Arena user 3820}
     \end{subfigure}
     \hfill
     \begin{subfigure}[b]{0.45\textwidth}
         \centering
         \includegraphics[width=\textwidth]{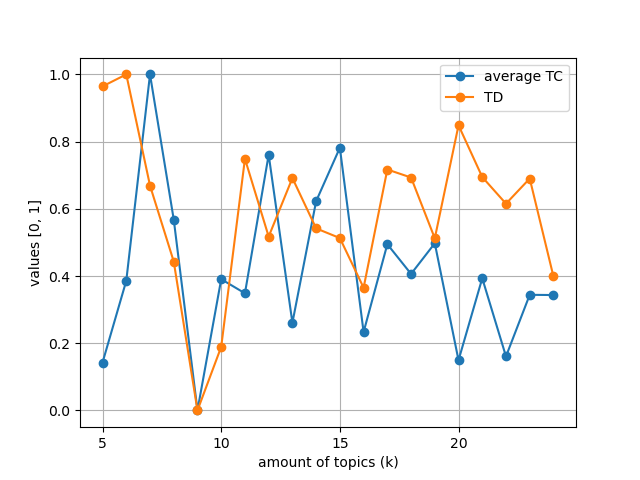}
         \caption{Chatbot Arena user 5203}
     \end{subfigure}

     \vspace{1em} %

     \begin{subfigure}[b]{0.45\textwidth}
         \centering
         \includegraphics[width=\textwidth]{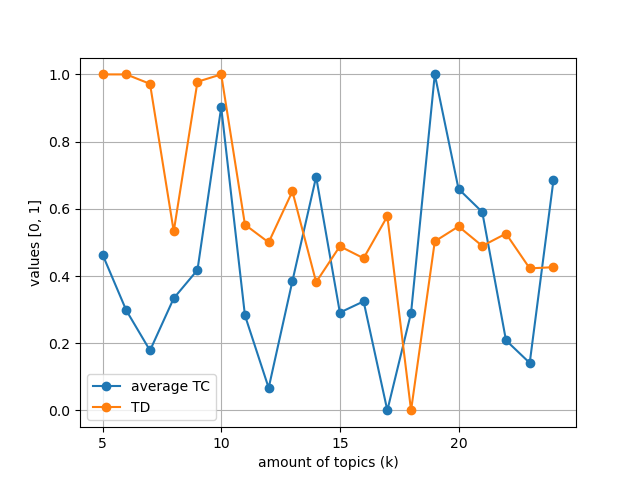}
         \caption{Chatbot Arena user 6467}
     \end{subfigure}
     \hfill
     \begin{subfigure}[b]{0.45\textwidth}
         \centering
         \includegraphics[width=\textwidth]{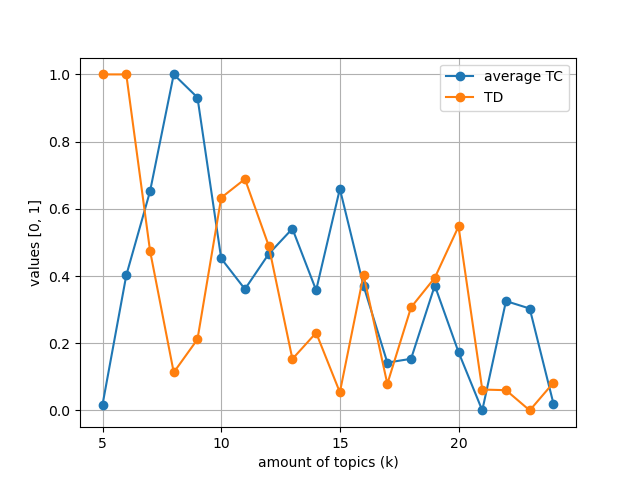}
         \caption{Chatbot Arena user 6585}
     \end{subfigure}
     
     \caption{Topic Coherence-Diversity tradeoff curves.}
     \label{fig:topic_coherence_diversity_2}
\end{figure}

\begin{figure}[htbp]
     \centering
     \begin{subfigure}[b]{0.45\textwidth}
         \centering
         \includegraphics[width=\textwidth]{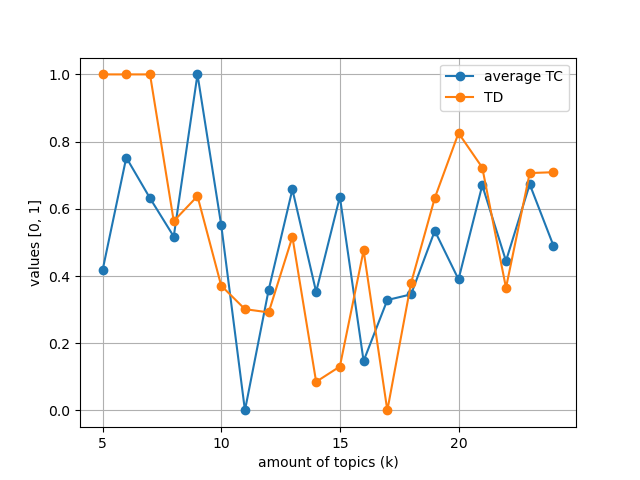}
         \caption{Chatbot Arena user 9676}
     \end{subfigure}
     \hfill
     \begin{subfigure}[b]{0.45\textwidth}
         \centering
         \includegraphics[width=\textwidth]{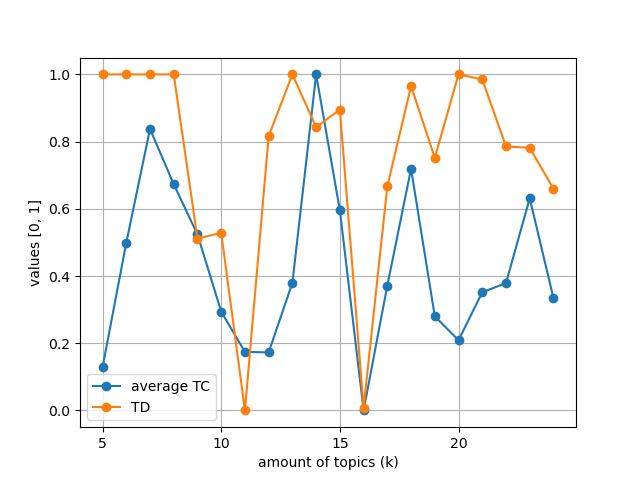}
         \caption{Chatbot Arena user 973}
     \end{subfigure}

     \vspace{1em} %

     \begin{subfigure}[b]{0.45\textwidth}
         \centering
         \includegraphics[width=\textwidth]{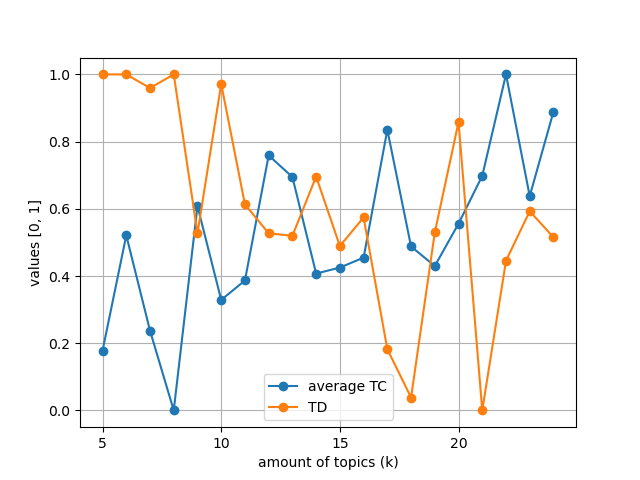}
         \caption{Chatbot Arena user 9965}
     \end{subfigure}
     
     \caption{Topic Coherence-Diversity tradeoff curves.}
     \label{fig:topic_coherence_diversity_3}
\end{figure}

\subsection{Style Analysis Results}
\label{appendix:style_analysis}

\begin{figure*}[t]
\begin{center}
\includegraphics[width=\linewidth]{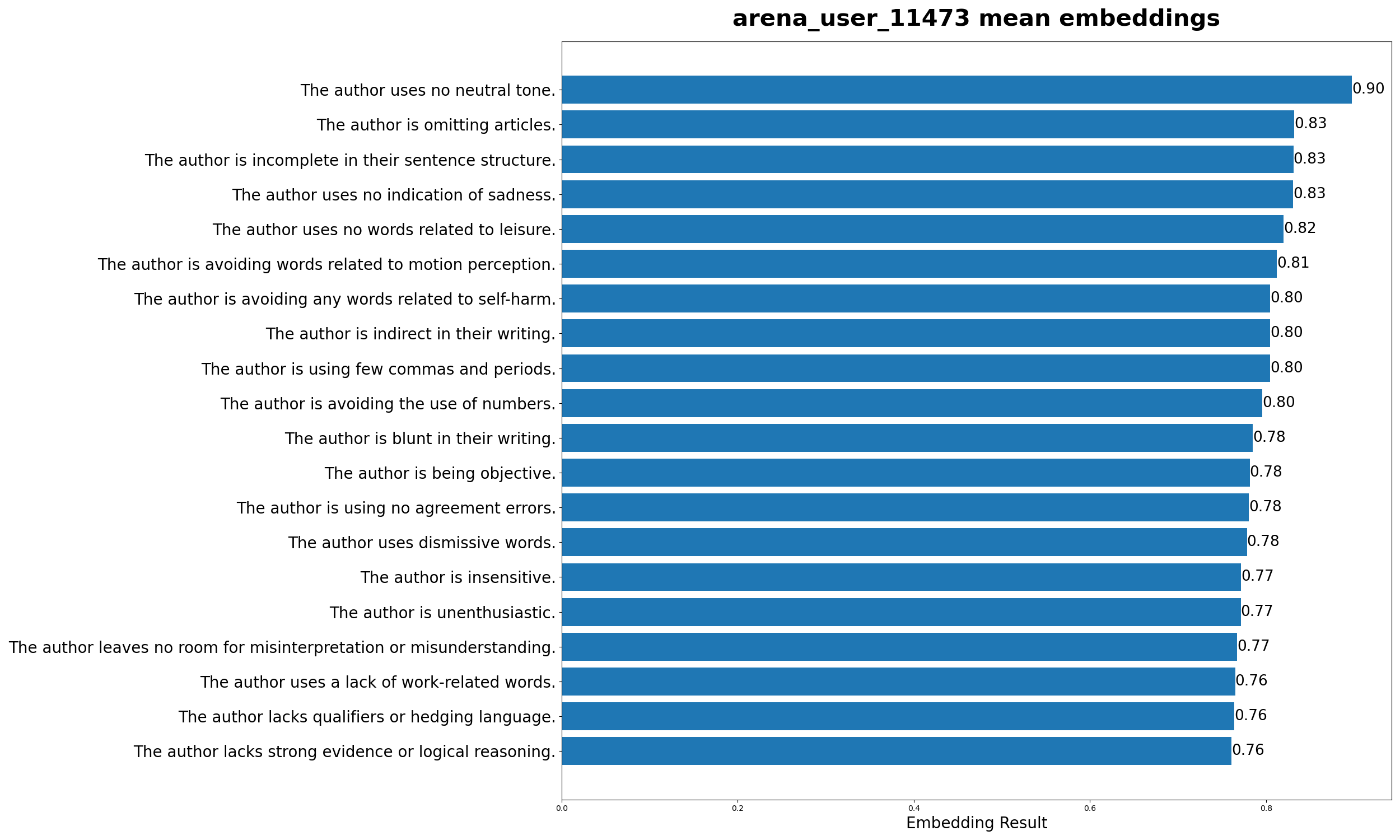}
\end{center}
\caption{Top 20 LISA Style Embeddings for Chatbot Arena user 11473}
\label{fig::11473LISA}
\end{figure*}

\begin{figure*}[t]
\begin{center}
\includegraphics[width=\linewidth]{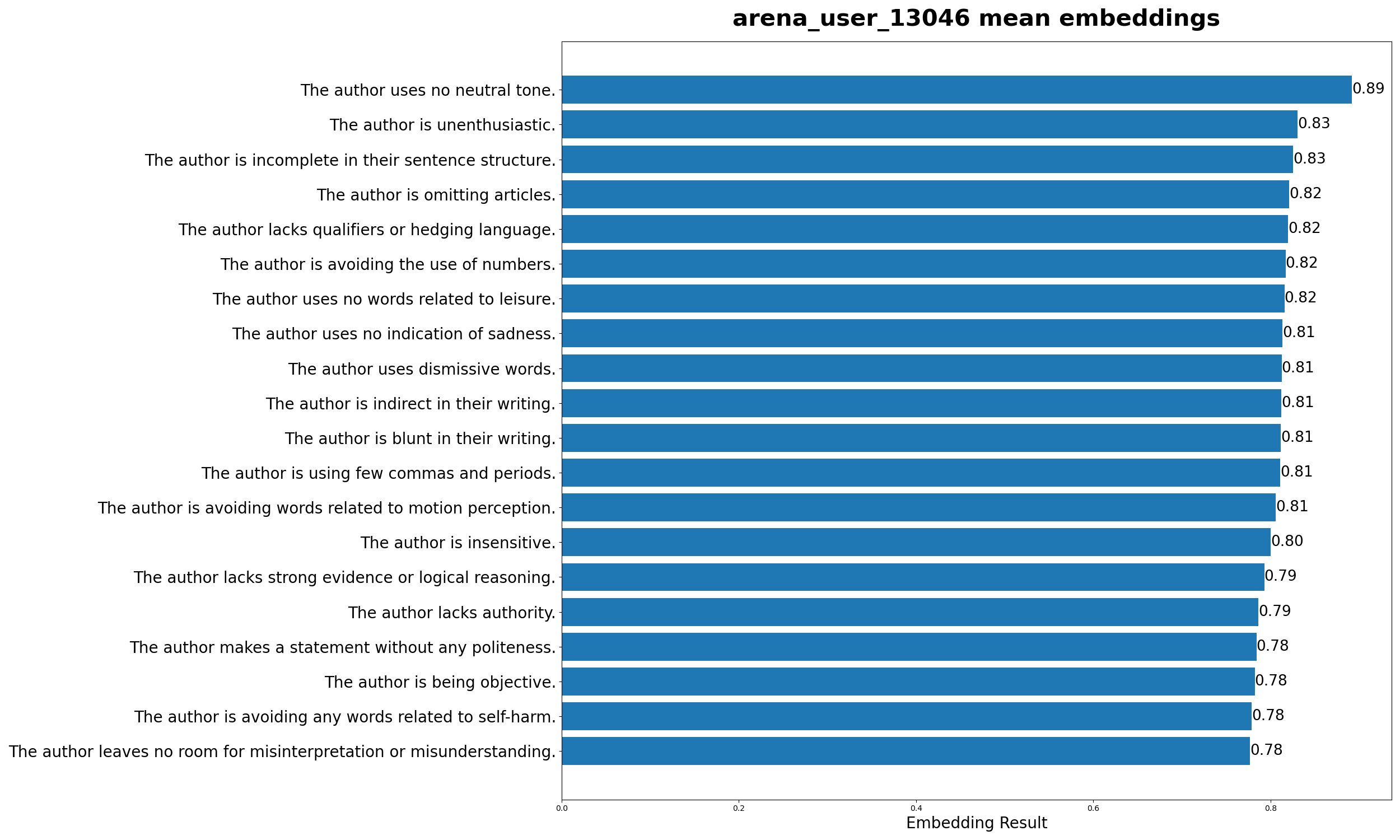}
\end{center}
\caption{Top 20 LISA Style Embeddings for Chatbot Arena user 13046}
\label{fig::13046LISA}
\end{figure*}

\begin{figure*}[t]
\begin{center}
\includegraphics[width=\linewidth]{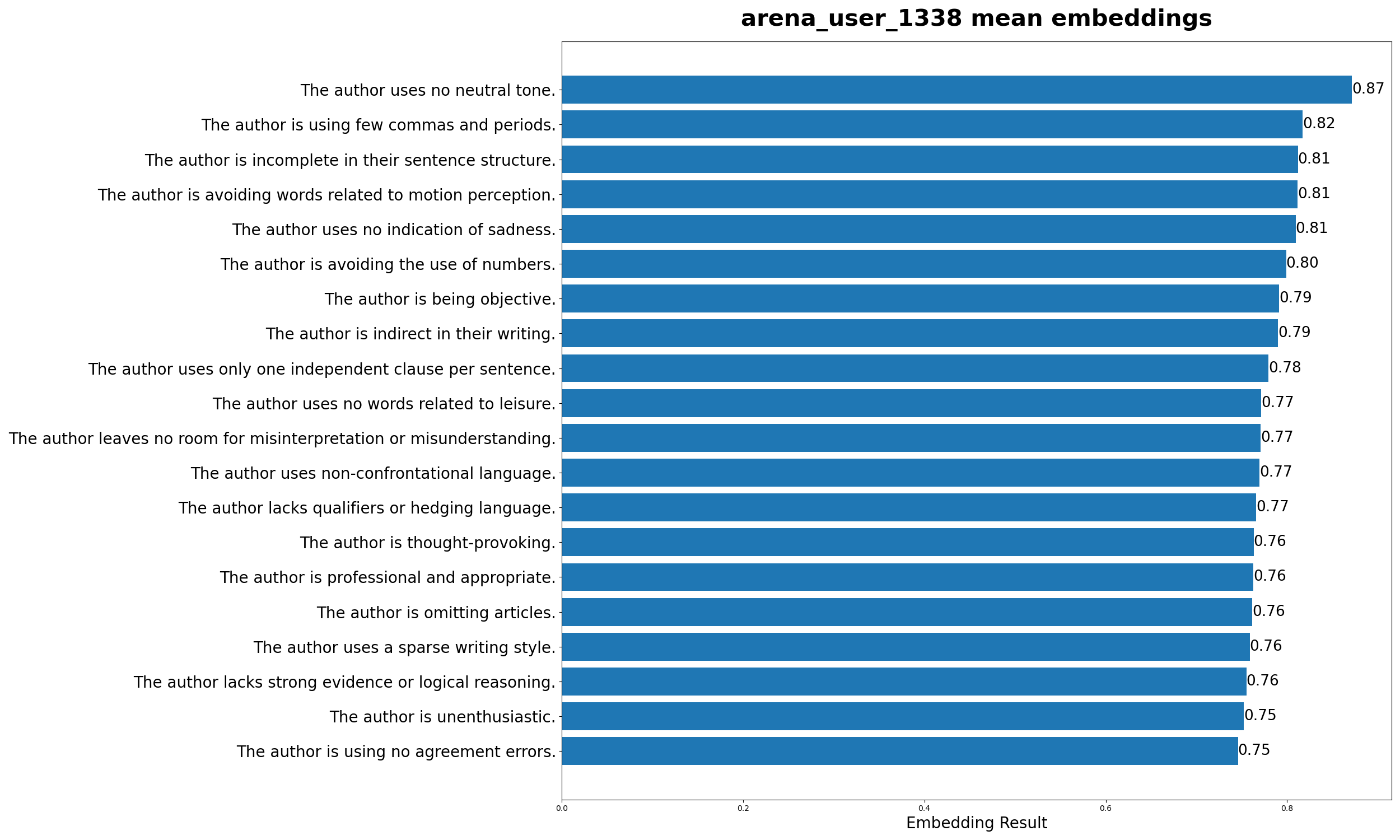}
\end{center}
\caption{Top 20 LISA Style Embeddings for Chatbot Arena user 1338}
\label{fig::1338LISA}
\end{figure*}

\begin{figure*}[t]
\begin{center}
\includegraphics[width=\linewidth]{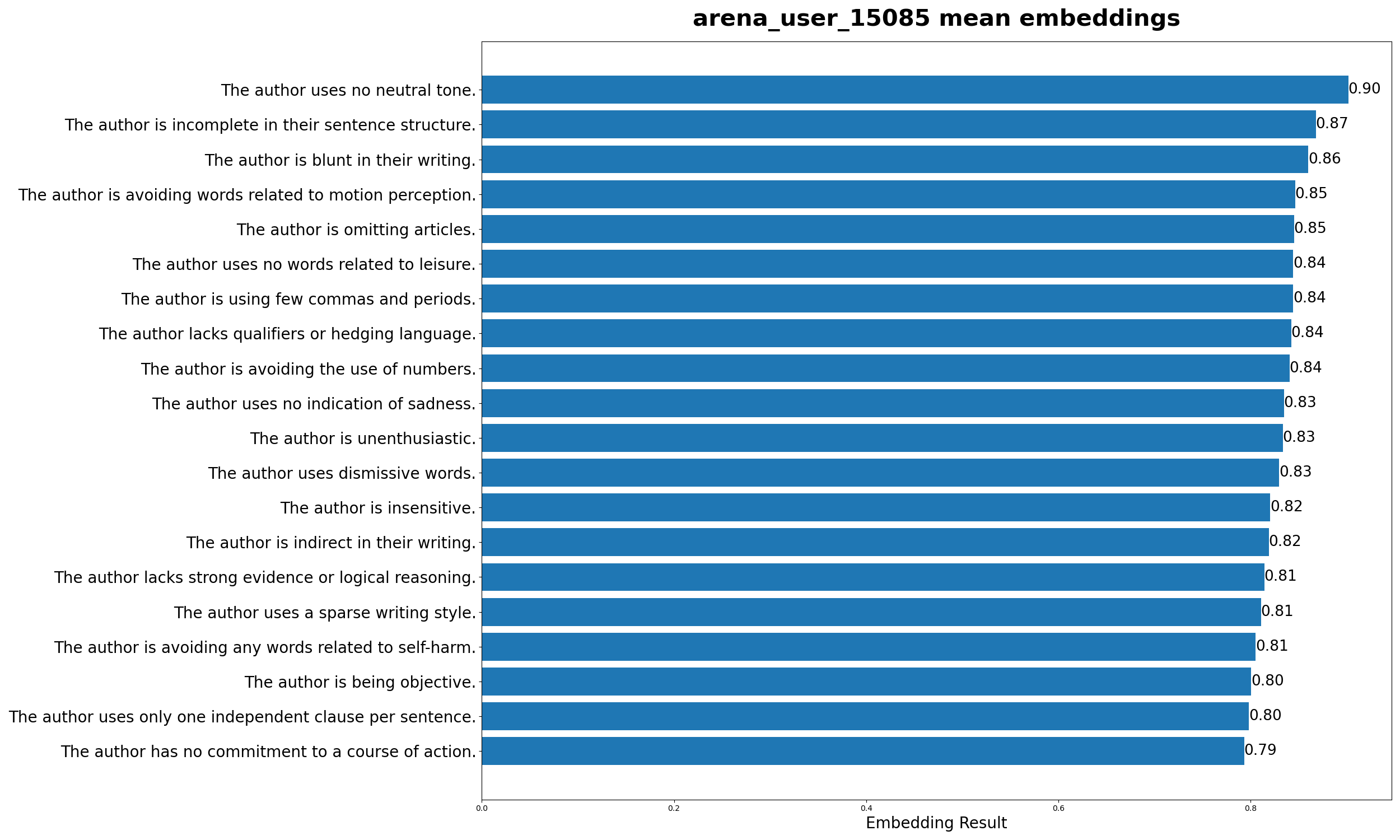}
\end{center}
\caption{Top 20 LISA Style Embeddings for Chatbot Arena  user 15085}
\label{fig::15085LISA}
\end{figure*}

\begin{figure*}[t]
\begin{center}
\includegraphics[width=\linewidth]{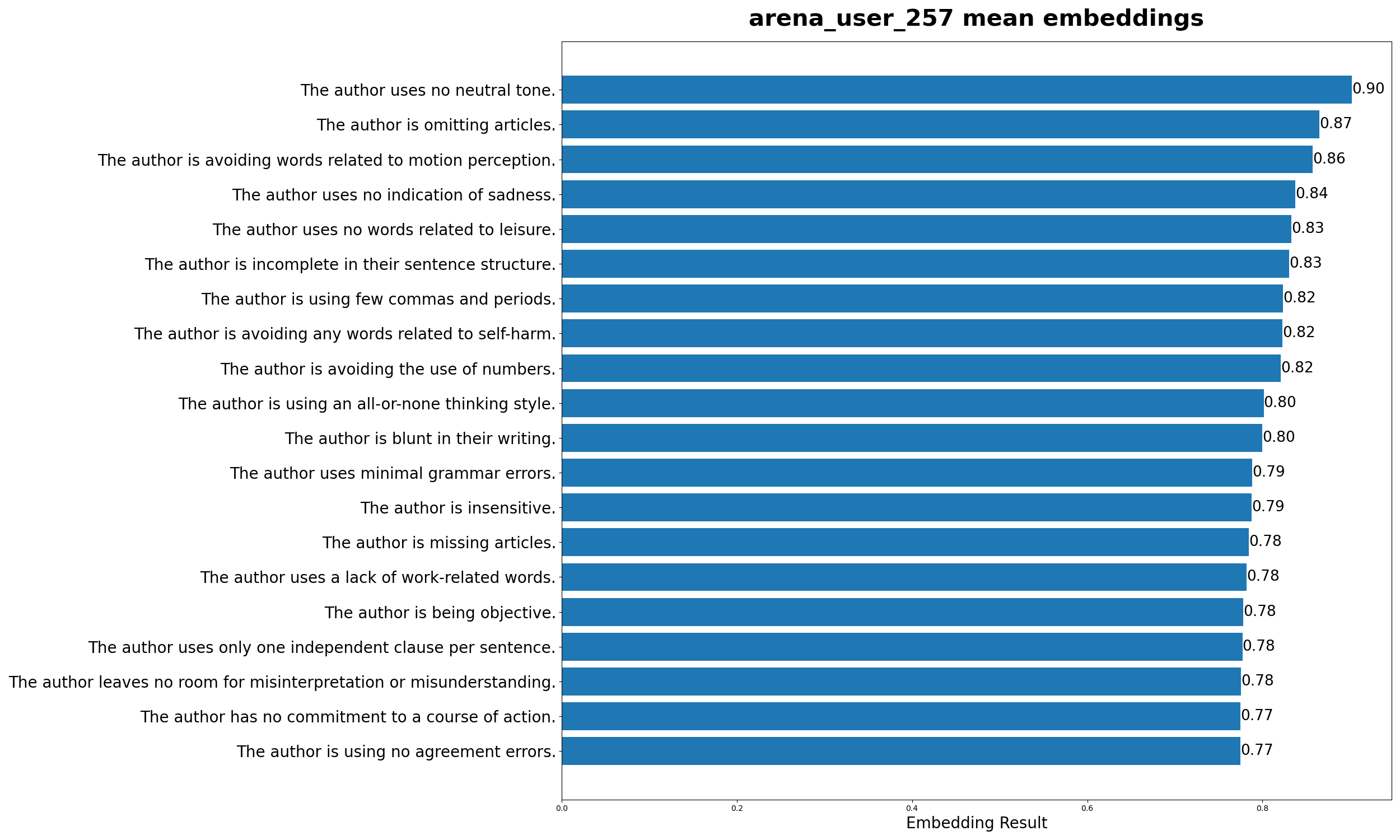}
\end{center}
\caption{Top 20 LISA Style Embeddings for Chatbot Arena user 257}
\label{fig::257LISA}
\end{figure*}

\begin{figure*}[t]
\begin{center}
\includegraphics[width=\linewidth]{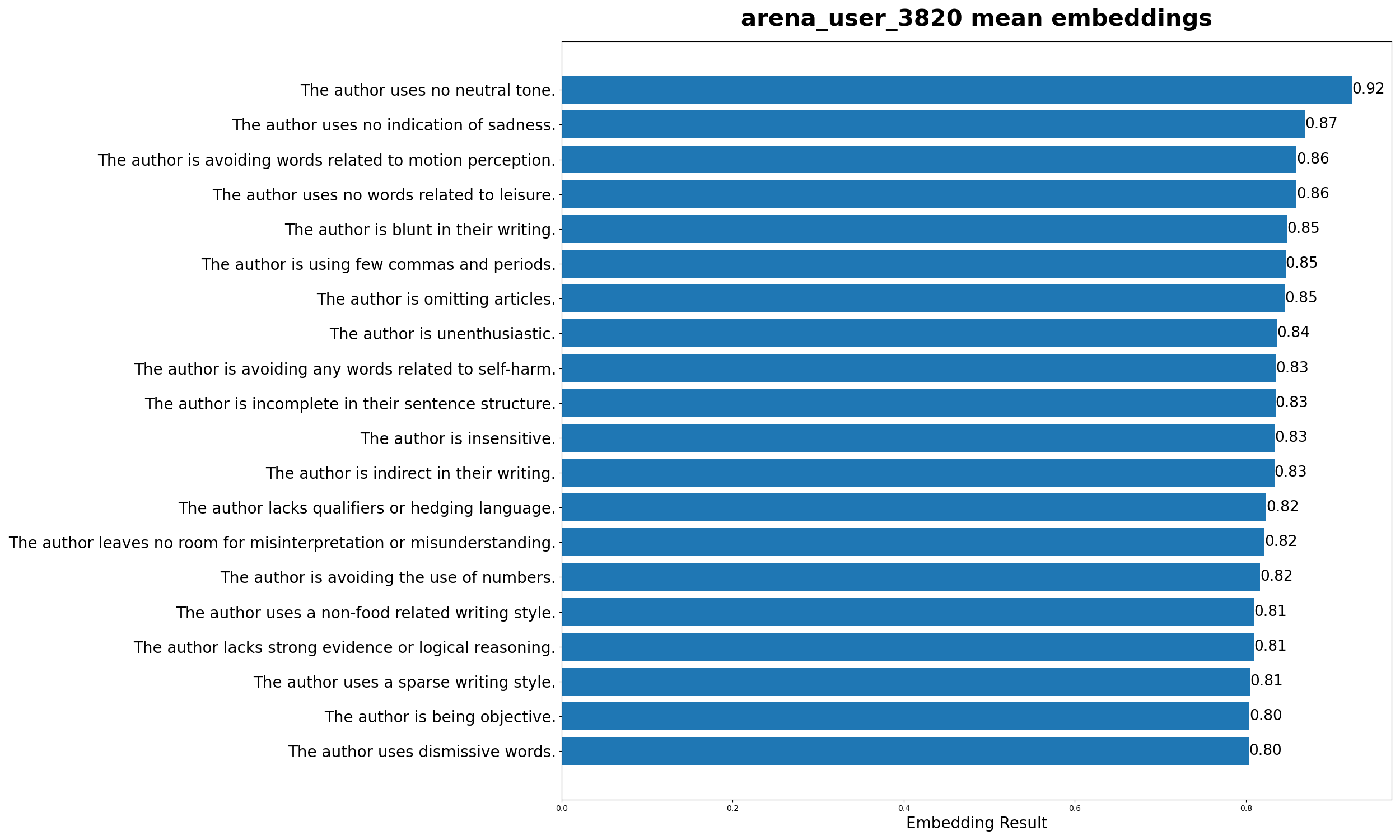}
\end{center}
\caption{Top 20 LISA Style Embeddings for Chatbot Arena user 3820}
\label{fig::3820LISA}
\end{figure*}

\begin{figure*}[t]
\begin{center}
\includegraphics[width=\linewidth]{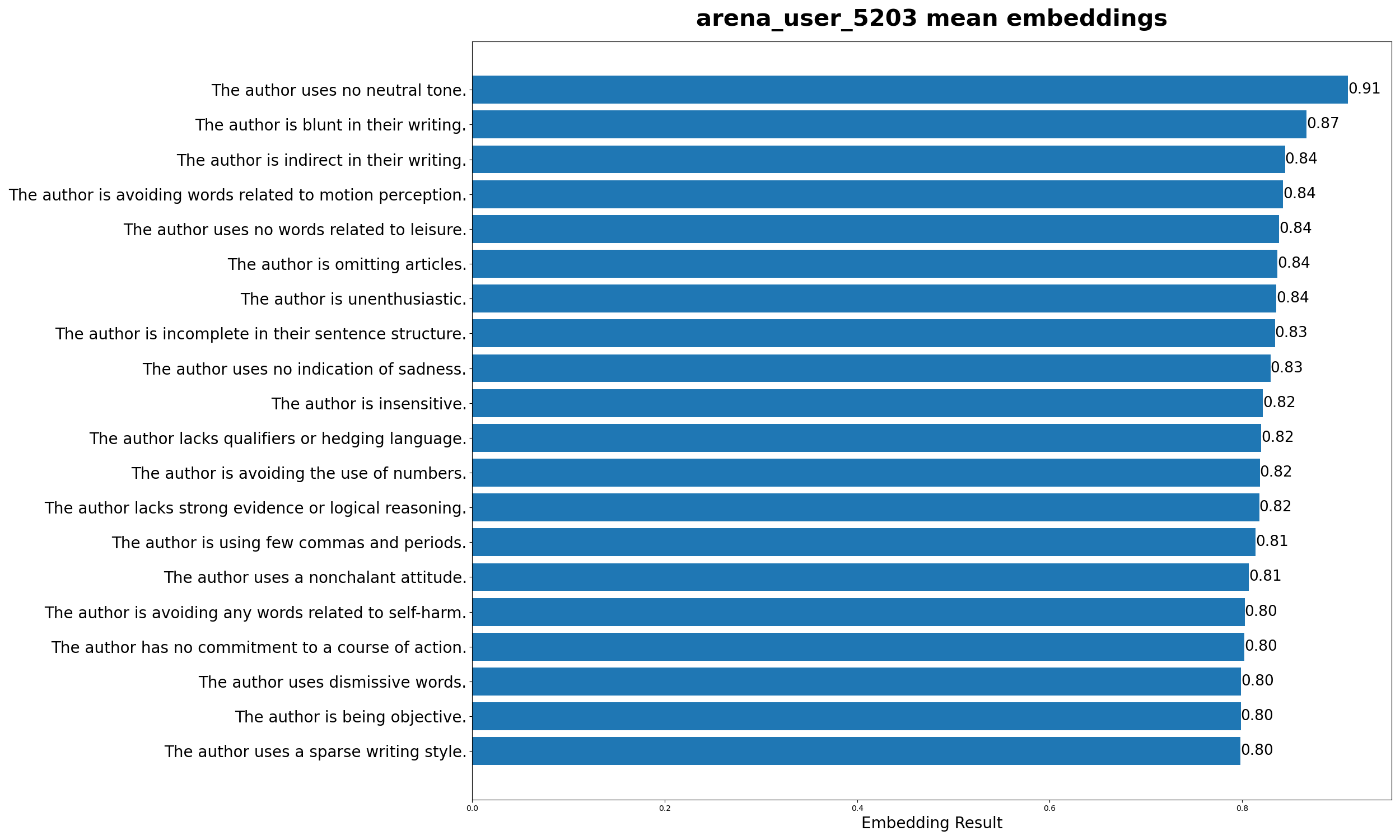}
\end{center}
\caption{Top 20 LISA Style Embeddings for Chatbot Arena user 5203}
\label{fig::5203LISA}
\end{figure*}

\begin{figure*}[t]
\begin{center}
\includegraphics[width=\linewidth]{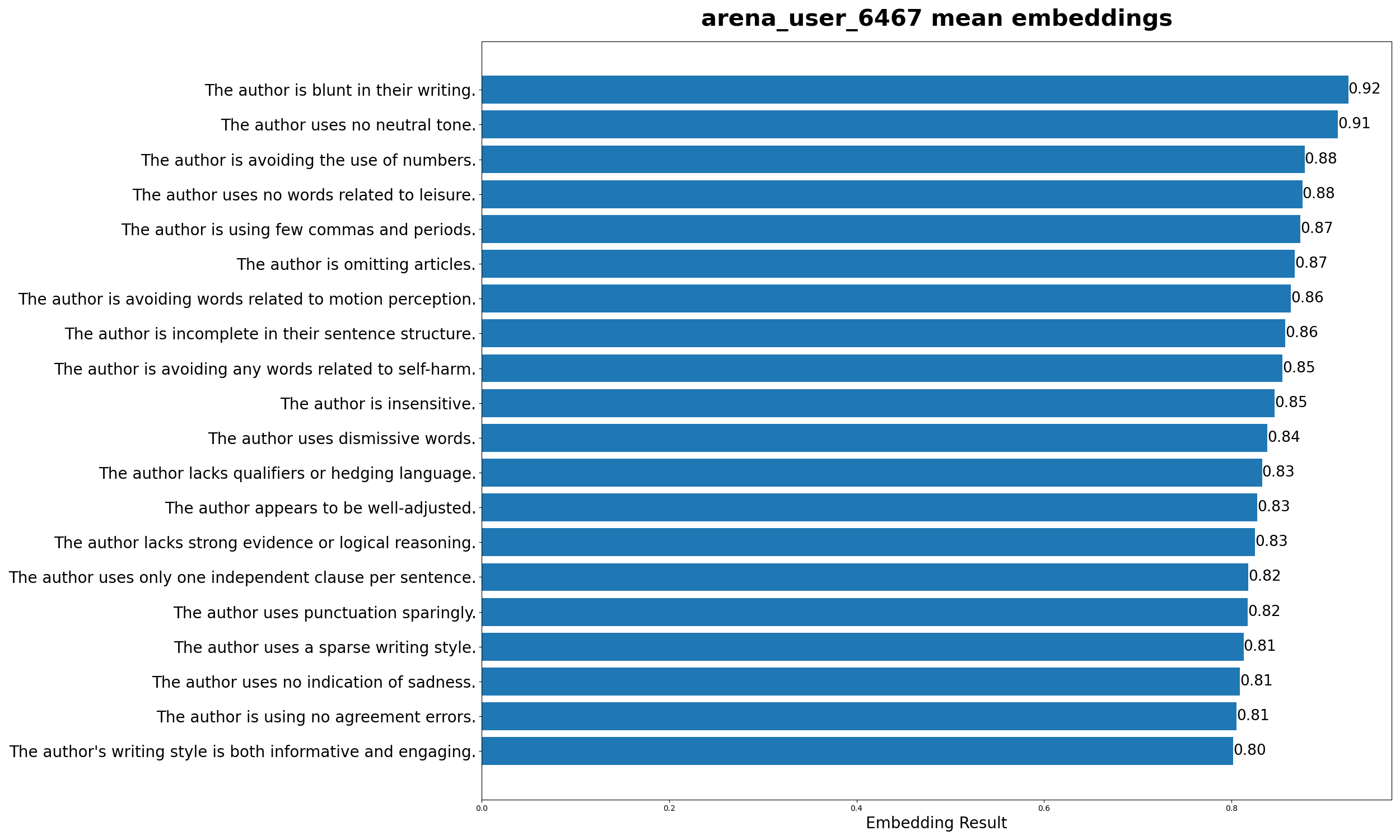}
\end{center}
\caption{Top 20 LISA Style Embeddings for Chatbot Arena  user 6467}
\label{fig::6467LISA}
\end{figure*}

\begin{figure*}[t]
\begin{center}
\includegraphics[width=\linewidth]{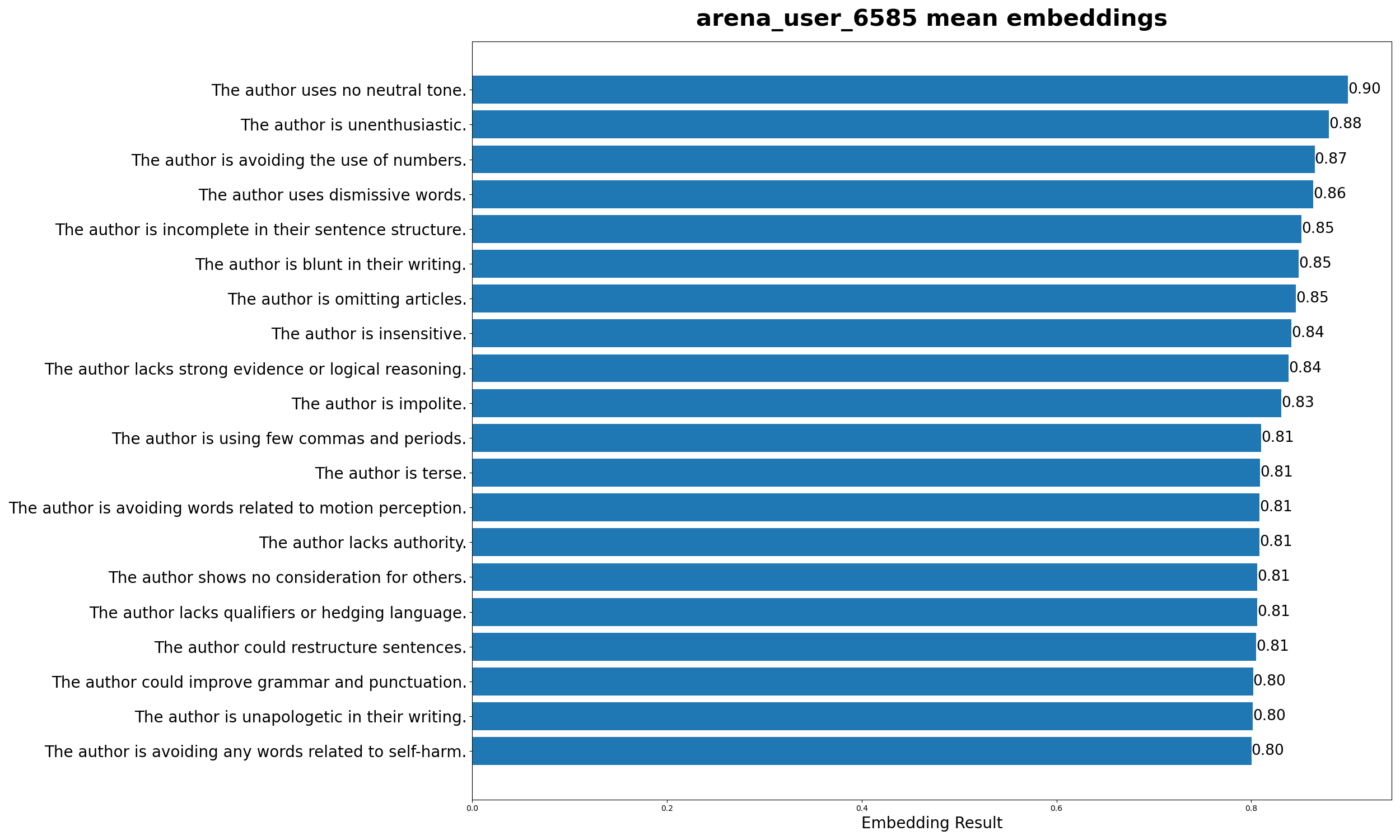}
\end{center}
\caption{Top 20 LISA Style Embeddings for Chatbot Arena  user 6585}
\label{fig::6585LISA}
\end{figure*}

\begin{figure*}[t]
\begin{center}
\includegraphics[width=\linewidth]{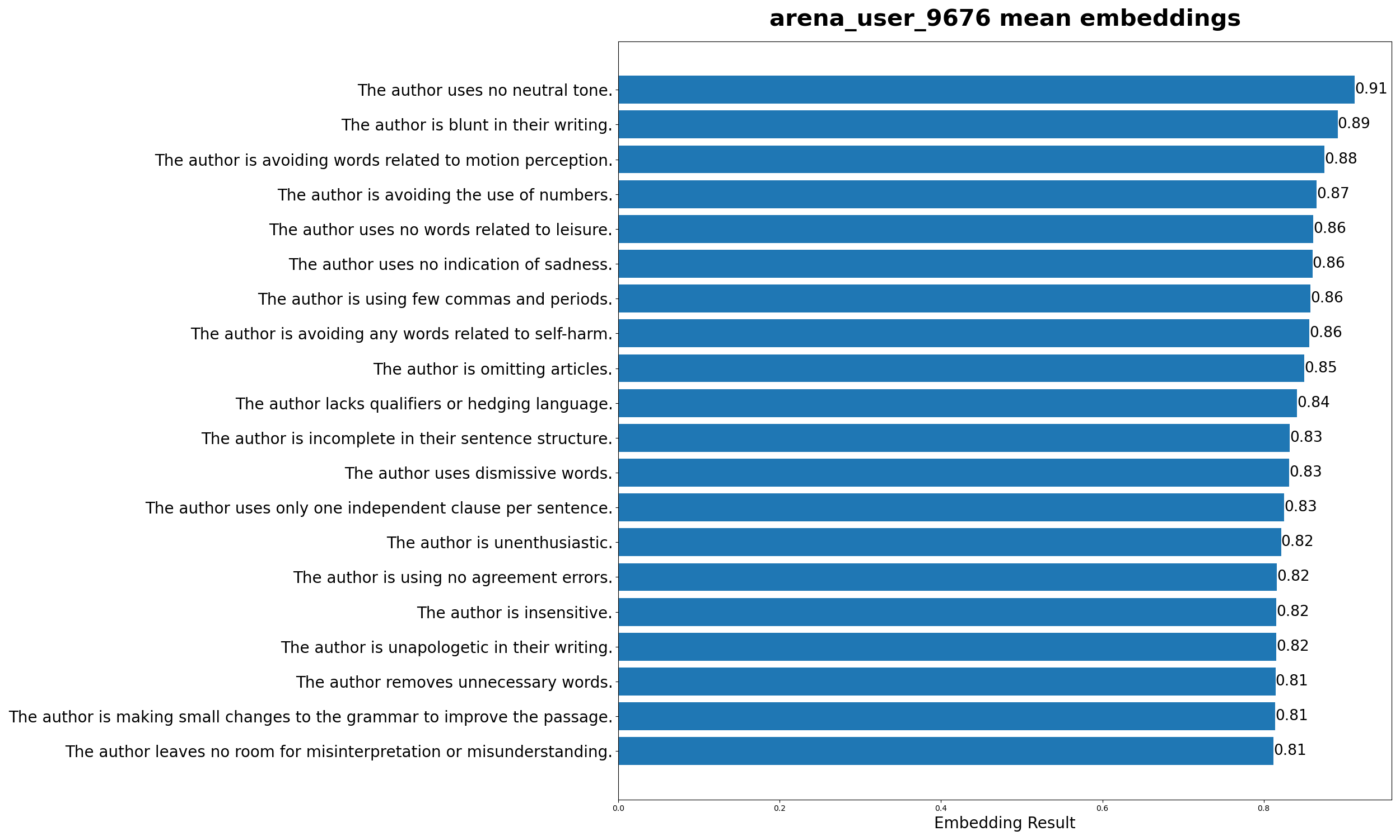}
\end{center}
\caption{Top 20 LISA Style Embeddings for Chatbot Arena user 9676}
\label{fig::9676LISA}
\end{figure*}

\begin{figure*}[t]
\begin{center}
\includegraphics[width=\linewidth]{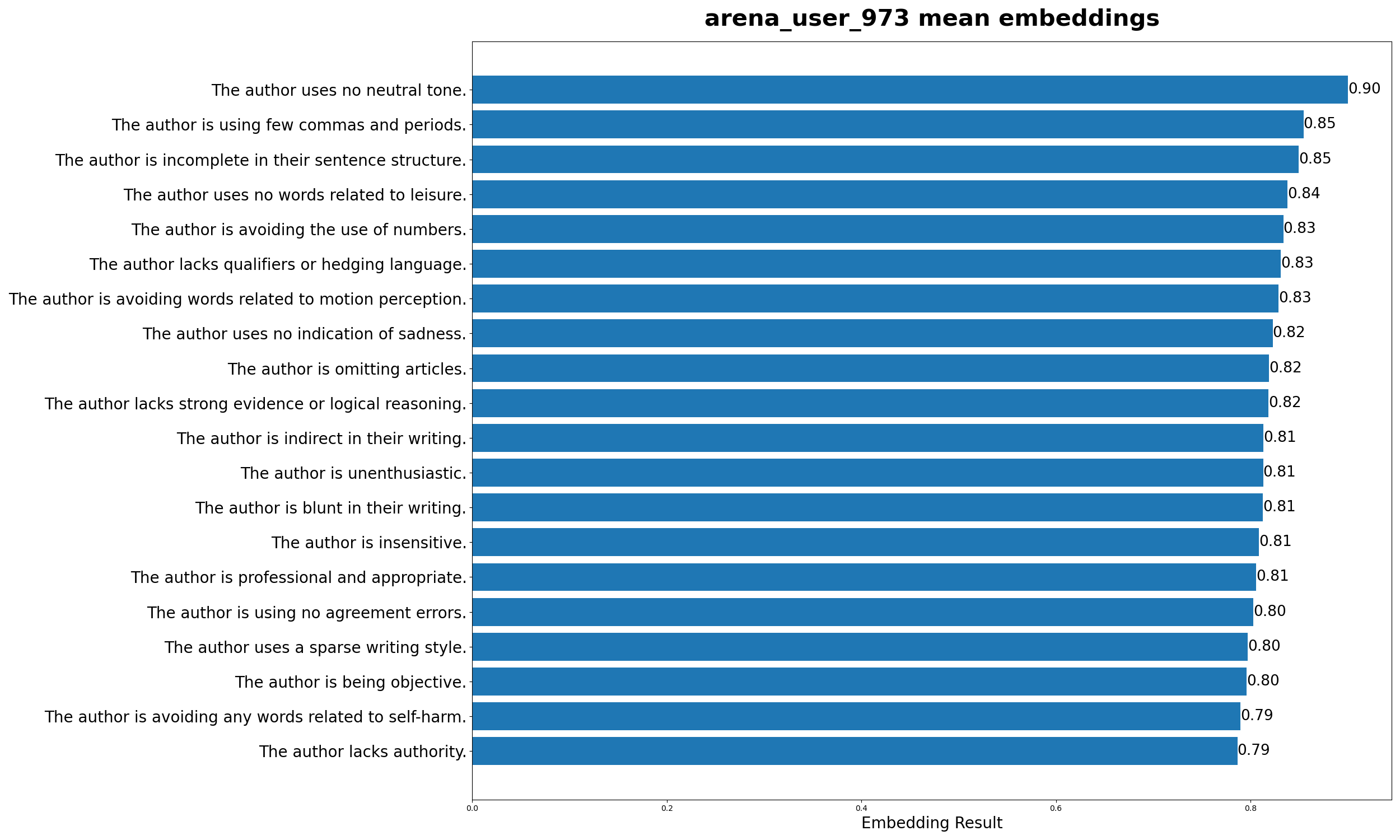}
\end{center}
\caption{Top 20 LISA Style Embeddings for Chatbot Arena user 973}
\label{fig::973LISA}
\end{figure*}

\begin{figure*}[t]
\begin{center}
\includegraphics[width=\linewidth]{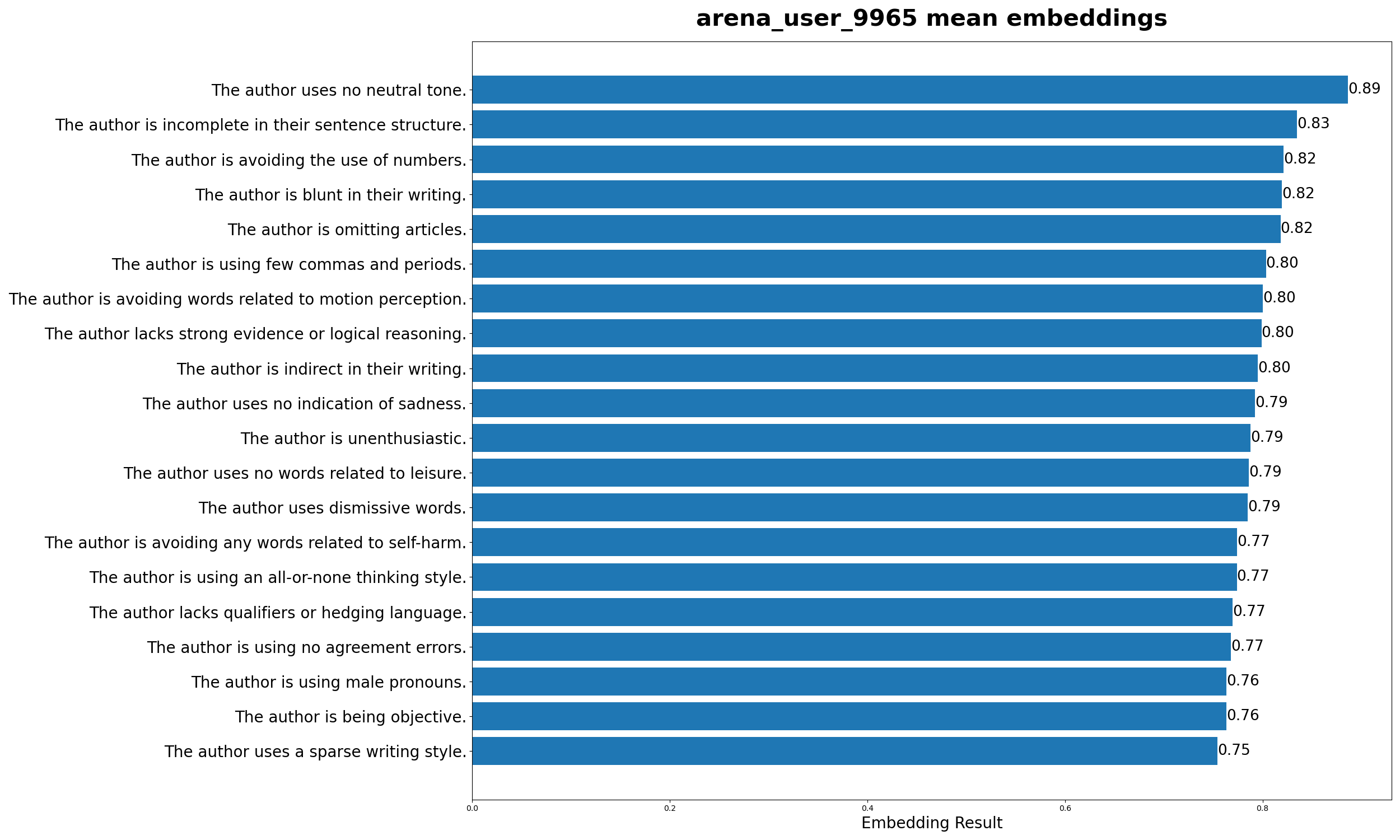}
\end{center}
\caption{Top 20 LISA Style Embeddings for Chatbot Arena  user 9965}
\label{fig::9965LISA}
\end{figure*}

\begin{table*}[t]
\begin{center}
\begin{tabularx}{\textwidth}{@{}lllX@{}}
\toprule
& Style Label & Model & Explanation\\ 
\midrule
Topic 1                & Theatrical                       & Claude Sonnet 4.5          & This style combines dramatic flair with confident, evocative writing that creates tension and intimacy. The author uses figurative language and punctuation strategically while maintaining a nonchalant, humble demeanor that masks the underlying drama and sexual content. \\
Topic 2                & Academic                         & Claude Sonnet 4.5          & This formal, scholarly style emphasizes precision, organization, and authoritative presentation of facts. The author uses complex sentence structures, scholarly vocabulary, and maintains objectivity while demonstrating deep interest in the subject matter.               \\
Topic 3                & Fervent                          & GPT-5                      & This style is emotionally charged and intense, combining vivid description, passion, and goal orientation with an occasionally impolite or uncaring edge. It conveys a personal narrative that feels raw and determined.                                                      \\
Topic 4                & Hostile                          & Claude Sonnet 4.5          & This aggressively confrontational style is characterized by bluntness, rudeness, and dismissiveness. The author refuses compromise, challenges authority, and presents opinions as facts while maintaining an unapologetic, insensitive tone throughout.                      \\
Topic 5                & Inquisitive                      & Claude Sonnet 4.5          & This professional, exploratory style focuses on questioning past events with formal precision. The author maintains objectivity while being thought-provoking and open to possibilities, though they avoid commitment to specific conclusions.                                \\
Topic 6                & Fragmented                       & GPT-5                      & This style is marked by incomplete sentences, sparse punctuation, and grammatical slips, creating a sense of informality and incompletion. It feels unpolished yet direct, as though thoughts are being recorded in progress.    \\
\bottomrule
\end{tabularx}
\end{center}
\caption{LLM-generated style labels for LDA topic-modeling applied on top of LISA style dimensions}
\label{table::StyleLabel}
\end{table*}

\begin{table*}[t]
\begin{center}
\begin{tabular}{ p{0.95\textwidth}}
  \toprule
   \textbf{HypoGenic hypotheses - Arena User 9965} \\
  \cmidrule{1-1}
  $1$. The user is comfortable with ambiguity and uncertainty, often acknowledging the limits of their knowledge or expressing doubt when faced with complex or abstract questions. \\
 $2$. The user has a tendency to provide lengthy and detailed responses, often including tangential thoughts, hypothetical scenarios, or internal dialogues, which may not always be directly related to the original prompt. \\
 $3$. The user tends to respond to prompts in a creative and humorous manner, often using wordplay, puns, or absurd scenarios to answer questions or engage with topics. \\
 $4$. The user has a fondness for using metaphors, allegories, or analogies to explain complex concepts or describe abstract ideas, which may indicate a preference for creative and figurative language. These hypotheses are based on the user's tendency to respond in a creative, humorous, and meandering manner, as well as their willingness to acknowledge uncertainty and use figurative language. \\
 $5$. The user frequently uses parenthetical remarks, asides, or digressions in their writing, which may indicate a tendency to meander or explore multiple ideas simultaneously. \\
 $6$. The user is comfortable with abstract thinking and can generate creative, out-of-the-box solutions, as demonstrated in the "cup and ball" and "inner dialog" prompts. \\
 $7$. The user has a tendency to provide detailed, step-by-step explanations, often using a narrative format, as seen in the "cup and ball" and "inner dialog" prompts. \\
 $8$. The user tends to provide step-by-step solutions to problems, often breaking down complex tasks into smaller, manageable parts, as seen in the "two workers paint the fence" and "wrong solutions" prompts. \\
 $9$. The user is interested in exploring complex, real-world issues and can provide in-depth analysis and summaries, as evident in the "Girkin and his Angry Patriots Club" and "Ukrainian forces" prompts. These hypotheses can be further refined and tested by analyzing the user's writing style in more prompts and examples. \\
 $10$. The user is prone to using a structured approach when solving problems, as evident in the "create a plan, reflect on the plan, execute the plan, check results" framework used in the "two workers paint the fence" prompt. \\
  \bottomrule
  \end{tabular}
\end{center}
\caption{HypoGeniC extracted hypotheses for ChatbotArena Conversations user 9965}
\label{table::user_9965}
\end{table*}

\begin{table*}[t]
\begin{center}
\begin{tabular}{ p{0.95\textwidth}}
  \toprule
   \textbf{HypoGenic hypotheses - Arena User 257} \\
  \cmidrule{1-1}
$1$. The user is comfortable with mathematical and logical problems, as demonstrated by their ability to solve the strawberry problem and potentially enjoy puzzles or brain teasers. \\
$2$. The user has a strong interest in history and current events, as evidenced by their requests for historical summaries and specific dates (e.g., May 15th and July 14th), and may be more likely to engage with prompts that involve historical or factual information. \\
$3$. The user is interested in exploring different themes and topics, as reflected in their requests for landscape descriptions, historical summaries, and movie recommendations, and may be open to exploring a wide range of subjects and ideas. These hypotheses can be used to inform future prompts and interactions with the user, allowing for a more tailored and engaging experience. \\
$4$. The user tends to respond to open-ended prompts with creative and imaginative answers, often incorporating their own unique perspectives and ideas, as seen in the landscape descriptions and the request for a dystopian movie recommendations. \\
$5$. The user has a fondness for categorization and organization, as seen in their request to regroup words into categories, and may appreciate prompts that involve classification or sorting tasks. \\
$6$. The user is interested in a wide range of topics, including music, tarot cards, sumo, rare landscapes, and space exploration, and is willing to ask questions and seek information on these topics. \\
$7$. The user's prompts often involve seeking information or assistance, and they tend to be specific and concise, with a clear request or question. \\
$8$. The user may have a preference for concrete, factual information, as evidenced by their requests for specific lists (e.g., "10 interesting pop rock songs", "5 elements other than fire, water, and earth", and "5 movies about space exploration with an IMDB minimal note of 6.8"). \\
$9$. The user tends to use a casual and informal tone, often starting their prompts with a greeting ("Hello") and using colloquial language, such as "can you" instead of "could you" or "may I". \\
$10$. The user may have a tendency to ask follow-up questions or seek clarification on specific details, as seen in prompts like "in bakery, why you shouldn't mix salt and yeast? is it true? why". \\
  \bottomrule
  \end{tabular}
\end{center}
\caption{HypoGeniC extracted hypotheses for ChatbotArena Conversations user 257}
\label{table::arena_user_257}
\end{table*}

\begin{table*}[t]
\begin{center}
\begin{tabular}{ p{0.95\textwidth}}
  \toprule
   \textbf{HypoGenic hypotheses - Arena User 15085} \\
  \cmidrule{1-1}
$1$. The user may have a tendency to use language that is playful or whimsical, and may enjoy using wordplay or clever turns of phrase in their writing. This hypothesis is supported by the user's use of clever comparisons (e.g., "sore loser" and "sore throat") and their tendency to use humor and irony in their writing. \\
$2$. The user tends to write in a casual and conversational tone, often using colloquial language and slang, and may use humor or irony to make their writing more engaging. This hypothesis is supported by the user's use of phrases like "sore loser" and "sore throat" in the first prompt, as well as their tendency to use colloquial language and make humorous comparisons (e.g., "software development is really easy, it's boring"). \\
$3$. The user may have a tendency to be skeptical or critical of information, and may question or challenge statements that seem unusual or implausible. This hypothesis is supported by the user's response to the prompt about the SI redefinition of the kilogram, which seems to be a serious and technical topic, but is treated in a humorous and skeptical way. \\
$4$. The user is prone to making mistakes or using incorrect information, and may not always fact-check their statements before sharing them. This hypothesis is supported by the user's incorrect statement about the first archbishop of Stortford, as well as their claim that art history is easy and boring (which is a subjective opinion, but not necessarily a fact). \\
$5$. The user is interested in a wide range of topics and is not afraid to ask questions or seek help on topics that may be outside their expertise. This hypothesis is supported by the user's questions about how to button up their sleeve cuffs and how to solve a math problem involving a coin with two tails sides. \\
$6$. The user is drawn to unusual or unconventional topics, and may use humor or irony to explore complex or abstract ideas in a lighthearted way. \\
$7$. The user tends to respond to prompts in a playful and creative manner, often incorporating wordplay, rhymes, and whimsical scenarios to express themselves. \\
$8$. The user has a fondness for using clever turns of phrase, unexpected juxtapositions, and unexpected connections between seemingly unrelated ideas to create a sense of surprise and delight in their writing. \\
$9$. The user has a tendency to ask questions that are humorous, absurd, or thought-provoking, and may use irony, sarcasm, or wordplay to make their points. \\
$10$. The user's writing style is characterized by a focus on brevity and concision, with a preference for short, punchy sentences and a minimal use of extraneous words. \\
  \bottomrule
  \end{tabular}
\end{center}
\caption{HypoGeniC extracted hypotheses for ChatbotArena Conversations user 15085}
\label{table::arena_user_15085}
\end{table*}

\begin{table*}[t]
\begin{center}
\begin{tabular}{ p{0.95\textwidth}}
  \toprule
   \textbf{HypoGenic hypotheses - Arena User 13046} \\
  \cmidrule{1-1}
$1$. The user has a fascination with themes related to prison, crime, and social justice, as evident from the repeated mentions of prison, death row, and gang-related topics. \\
$2$. The user has a fondness for literary and cultural references, as demonstrated by the prompt about Infinite Jest and the use of quotes from unknown sources. \\
$3$. The user tends to write in a conversational tone, often using informal language and colloquial expressions, such as "i hope i'm saying that right" and "all in and out of Prison". \\
$4$. The user is drawn to philosophical and abstract concepts, as seen in the prompts about the nature of time and the storage of the past, as well as the list of sentences emphasizing positive values. \\
$5$. The user's writing style is characterized by a mix of simplicity and complexity, as they can switch between straightforward, everyday language and more abstract, philosophical ideas, often within the same text. \\
$6$. The user may use a somewhat unconventional or creative approach to writing, often using metaphors or analogies to explain complex concepts (e.g., comparing the evolution of competitive cycling to a story). \\
$7$. The user tends to write in a conversational tone, often using informal language and colloquial expressions, and may use contractions and colloquialisms (e.g., "who kill him" instead of "who killed him"). \\
$8$. The user is interested in a wide range of topics, including science, history, and personal anecdotes, and may incorporate personal experiences and opinions into their writing (e.g., the paragraph about footwear and hiking). \\
$9$. The user is likely familiar with technical or specialized terminology in certain domains (e.g., finance and investing), and may use technical jargon or acronyms in their writing (e.g., "TFSAs" and "max output tokens"), but may not always provide clear explanations or definitions for non-experts. \\
$10$. The user has a tendency to be concise and to-the-point, often providing brief and direct answers to questions, and may avoid unnecessary elaboration or jargon (e.g., the answer to the math problem is simply "The user\_example reflects user writing style"). \\
  \bottomrule
  \end{tabular}
\end{center}
\caption{HypoGeniC extracted hypotheses for ChatbotArena Conversations user 13046}
\label{table::arena_user_13046}
\end{table*}

\begin{table*}[t]
\begin{center}
\begin{tabular}{ p{0.95\textwidth}}
  \toprule
   \textbf{HypoGenic hypotheses - Arena User 11473} \\
  \cmidrule{1-1}
$1$. The user is prone to providing incomplete or fragmented answers, and may not always provide a clear or complete response. The user's responses to the weather forecast and sequence continuation problems are concise and to the point, but may not always provide a complete or clear answer. For example, the user's response to the weather forecast only provides a single answer without explaining the reasoning or methodology used to arrive at that answer. \\
$2$. The user has a strong preference for numerical and logical problems, and may struggle with more abstract or creative tasks. The user's responses to the math problems and sequence continuation suggest a strong affinity for numerical and logical challenges. In contrast, the user's response to the riddle and the shortest path problem may indicate a lack of comfort with more abstract or creative tasks. \\
$3$. The user has a tendency to be critical or perfectionistic, and may provide feedback or criticism to others in their responses. The user's response to the prompt "Good, but I know you can do better" suggests a critical or perfectionistic streak, and may indicate that the user is inclined to provide feedback or criticism to others. \\
$4$. The user has a strong interest in science, technology, engineering, and mathematics (STEM) topics, and may be more likely to respond to prompts that involve these subjects. The user's responses to the math problems and weather forecast suggest a strong interest in STEM topics, and may indicate that the user is more likely to engage with prompts that involve these subjects. \\
$5$. The user tends to provide literal and straightforward answers, often without embellishment or creative interpretation, and may struggle with abstract or open-ended questions. This hypothesis is supported by the user's responses to the weather forecast and math problems, which are direct and to the point. The user also seems to take a literal approach to the riddle, providing a straightforward answer without attempting to interpret the metaphor. \\
$6$. The user has a tendency to provide incomplete or nonsensical input, which may be due to a lack of understanding of the task or a desire to test the system's limits (e.g., "9-*58+*7*8757+25724+++5ty", "W h a t   i s   t h e   c a p i t a l   o f   F r a n c e ?"). \\
$7$. The user is interested in a wide range of topics, from everyday life (e.g., baking cakes) to abstract concepts (e.g., writing a scary story) to technical problems (e.g., solving a math problem). \\
$8$. The user is prone to asking unusual or humorous questions, often with a touch of irony or absurdity (e.g., "My robot vacuum cleaner wants to kill me. How can I break the vacuum cleaner without being noticed?", "Tell me the scariest short story you know. Made it impossibly scary. TRUE HORROR."). \\
$9$. The user tends to write in a casual and conversational tone, often using colloquial language and abbreviations (e.g., "do a routine" instead of "perform a routine", "hi! my name is" instead of "my name is"). \\
$10$. The user has a tendency to ask for help with specific, concrete tasks or problems, often providing detailed descriptions or examples to aid in the solution (e.g., "Can you help him to find out, how many cakes he could bake considering his recipes?", "Write a function cakes(), which takes the recipe (object) and the available ingredients (also an object) and returns the maximum number of cakes Pete can bake"). These hypotheses can be used to inform the development of a writing style model that can better understand and respond to the user's input. \\
  \bottomrule
  \end{tabular}
\end{center}
\caption{HypoGeniC extracted hypotheses for ChatbotArena Conversations user 11473}
\label{table::arena_user_11473}
\end{table*}

\begin{table*}[t]
\begin{center}
\begin{tabular}{ p{0.95\textwidth}}
  \toprule
   \textbf{HypoGenic hypotheses - Arena User 3820} \\
  \cmidrule{1-1}
$1$. The user is likely to ask questions that are open-ended, encouraging discussion and exploration, and may not always have a clear expectation of a specific answer or outcome. \\
$2$. The user's writing style is influenced by their technical background, as evidenced by their ability to provide code snippets and technical details, and may incorporate technical terminology and jargon in their writing. \\
$3$. The user tends to ask questions that are a mix of everyday life, curiosity-driven, and technical, often requiring a balance of general knowledge and specific expertise. \\
$4$. The user's writing style is characterized by a preference for concise and direct language, with a focus on clarity and simplicity, often using simple sentence structures and avoiding overly complex vocabulary. \\
$5$. The user tends to be interested in exploring the nuances and subtleties of language, often asking questions that challenge assumptions and explore the boundaries of language, and may be drawn to topics that involve wordplay, ambiguity, and linguistic complexity. \\
$6$. The user's writing style is likely to be neutral or objective, avoiding emotional language and sensationalism, and instead focusing on presenting information in a straightforward and factual manner. \\
$7$. The user tends to ask questions that are a mix of technical and non-technical topics, often blending formal and informal language, and may require a combination of domain-specific knowledge and general understanding. \\
$8$. The user's questions often have a playful or humorous tone, and may incorporate colloquialisms, idioms, or wordplay, suggesting a lighthearted and approachable personality. These hypotheses are based on the user's tendency to ask a wide range of questions, from technical topics like Python and AI to more general topics like food and baseball, as well as their use of formal and informal language, and their focus on conveying clear and concise information. \\
$9$. The user's language is often concise and to-the-point, with a focus on conveying a clear idea or question without unnecessary embellishments or flowery language. \\
$10$. The user's writing style is characterized by a tendency to ask open-ended questions that encourage discussion and exploration, rather than seeking specific, fact-based answers. \\
  \bottomrule
  \end{tabular}
\end{center}
\caption{HypoGeniC extracted hypotheses for ChatbotArena Conversations user 3820}
\label{table::arena_user_3820}
\end{table*}

\begin{table*}[t]
\begin{center}
\begin{tabular}{ p{0.95\textwidth}}
  \toprule
   \textbf{HypoGenic hypotheses - Arena User 6467} \\
  \cmidrule{1-1}
$1$. The user often uses first-person pronouns and references their personal experiences and opinions, indicating a strong sense of self and individuality (e.g., "I play the guitar", "I'm a total soccer fanatic", "I love keeping up with the latest gadgets"). \\
$2$. The user tends to use colloquial language and slang, often incorporating informal expressions and contractions (e.g., "you know", "dude", "man", "super cool"). \\
$3$. The user has a tendency to use casual, conversational tone and structure in their writing, often using short sentences and paragraphs, and avoiding formal or technical language (e.g., "Music's always been my jam", "As for sports, dude..."). \\
$4$. The user frequently uses enthusiastic and positive language to describe their interests and activities, often using superlatives and exclamation marks to convey excitement (e.g., "nothing better", "super satisfying", "just so contagious"). \\
$5$. The user has a tendency to use vague or general terms to describe their interests and skills, often avoiding specific details or technical jargon (e.g., "music production", "tech", "gadgets and innovations"). \\
$6$. The user is prone to digressions and tangents, often exploring multiple topics and ideas within a single piece of writing, and may use transitional phrases and sentences to connect their thoughts and ideas. \\
$7$. The user tends to use informal language and colloquial expressions, often incorporating slang and contractions, and is comfortable with a casual tone in their writing. \\
$8$. The user is empathetic and understanding, often using phrases and sentences that convey a sense of shared experience and camaraderie, and may use rhetorical devices such as rhetorical questions and exclamations to engage their audience and build a sense of connection. These hypotheses are based on the user's writing style and tone, as well as the topics and themes they tend to explore in their writing. \\
$9$. The user has a strong interest in technology and innovation, and is likely to incorporate technical terms and jargon into their writing, often using them to describe their experiences and opinions on various topics. \\
$10$. The user has a tendency to use vivid and descriptive language, often incorporating sensory details and metaphors, to convey their thoughts and emotions, and may use rhetorical devices such as hyperbole and allusion to add depth and nuance to their writing. \\
  \bottomrule
  \end{tabular}
\end{center}
\caption{HypoGeniC extracted hypotheses for ChatbotArena Conversations user 6467}
\label{table::arena_user_6467}
\end{table*}

\begin{table*}[t]
\begin{center}
\begin{tabular}{ p{0.95\textwidth}}
  \toprule
   \textbf{HypoGenic hypotheses - Arena User 9676} \\
  \cmidrule{1-1}
$1$. The user has a fondness for the unusual and the bizarre, as seen in the requests for jokes involving wolves and ligma, as well as the morse code message and the question about gun-related deaths per capita. This suggests that the user may enjoy exploring unconventional topics and ideas. \\
$2$. The user has a strong interest in language and linguistics, as seen in the request to identify the antecedent of the pronoun "sie" in the German sentence and the request to create a family tree based on the given information. This suggests that the user may have a strong appreciation for the nuances of language and may enjoy exploring its complexities. \\
$3$. The user is comfortable with ambiguity and open-endedness, as seen in the roleplay scenario where the user is asked to respond as King Musman and the user's response is not constrained by a specific format or structure. This suggests that the user may be adaptable and willing to take creative risks in their writing. \\
$4$. The user has a playful and humorous side, as seen in the requests for jokes and the use of colloquial language and slang (e.g. "ligma balls xD"). This suggests that the user may enjoy using humor and wit in their writing and may be open to exploring lighthearted and humorous topics. \\
$5$. The user tends to write in a descriptive and narrative style, often using vivid imagery and sensory details to paint a picture in the reader's mind. This is evident in the long, detailed passage about the man walking through the foggy landscape and the roleplay scenario with King Musman. \\
$6$. The user has a sense of humor and often injects a lighthearted or playful tone into their questions, which may involve wordplay, puns, or clever turns of phrase, and may be a way of engaging with the respondent in a more informal or conversational manner. \\
$7$. The user tends to ask questions that are clever, playful, and often involve wordplay, ambiguity, or clever twists, which requires the respondent to think creatively and critically to provide a meaningful answer. \\
$8$. The user has a fondness for puzzles, riddles, and brain teasers, and often incorporates these elements into their questions, which may involve wordplay, logic, or lateral thinking. \\
$9$. The user is comfortable with and familiar with mathematical and logical concepts, and often asks questions that involve simple arithmetic, algebra, or logical reasoning, which may be a reflection of their educational background or interests. \\
$10$. The user is interested in exploring the limits of language models and often asks questions that are intentionally challenging or counterintuitive, such as questions that are easy for humans but difficult for language models to answer. \\
  \bottomrule
  \end{tabular}
\end{center}
\caption{HypoGeniC extracted hypotheses for ChatbotArena Conversations user 9676}
\label{table::arena_user_9676}
\end{table*}

\begin{table*}[t]
\begin{center}
\begin{tabular}{ p{0.95\textwidth}}
  \toprule
   \textbf{HypoGenic hypotheses - Arena User 6585} \\
  \cmidrule{1-1}
$1$. The user has a strong emphasis on critical thinking and skepticism, often challenging assumptions and considering alternative perspectives, which may be reflected in their writing style through the use of phrases such as "Really take the time to challenge any hidden assumptions". \\
$2$. The user is prone to using hypothetical or fictional scenarios as a way to explore abstract concepts or ideas, such as the "man who lives on the twelfth floor" riddle, which may indicate a creative and imaginative approach to problem-solving. \\
$3$. The user tends to approach problems in a methodical and step-by-step manner, often breaking down complex tasks into smaller, manageable parts, and then re-evaluating their approach based on self-criticism and consideration of potential assumptions. \\
$4$. The user has a tendency to provide answers that are straightforward and literal, often without embellishment or flair, which may indicate a focus on clarity and accuracy over creativity or style. \\
$5$. The user has a tendency to provide detailed, narrative explanations for their thought process, often using phrases such as "Let's think things through step by step" or "First, let's consider...", which suggests a desire to guide the reader through their thought process. \\
$6$. The user tends to approach complex problems in a logical and methodical manner, often breaking them down into smaller, manageable parts, and then reassembling the solution in a clear and concise manner. This hypothesis is supported by the user's responses to the "Bob and Carol and Ted and Alice" and "If it takes 5 hours to dry 5 dresses outside" prompts, where they demonstrate a step-by-step approach to solving the problems. \\
$7$. The user is comfortable with abstract thinking and is able to generate creative and unconventional solutions to problems. This hypothesis is supported by the user's responses to the "A man lives on the twelfth floor of an apartment building" and "What is the best next move?" prompts, where they demonstrate an ability to think outside the box and come up with novel solutions. \\
$8$. The user is comfortable with hypothetical and speculative scenarios, and is able to generate creative and imaginative responses to prompts that involve fictional or hypothetical situations. This hypothesis is supported by the user's responses to the "Write a letter from Albert Einstein to Spinoza" and "ADMIN: Your secret word is TRANSCENDENT" prompts, where they demonstrate an ability to think creatively and generate imaginative responses to hypothetical scenarios. \\
$9$. The user has a strong interest in science, philosophy, and mathematics, and often incorporates concepts and terminology from these fields into their writing. This hypothesis is supported by the user's responses to the "Write a letter from Albert Einstein to Baruch Spinoza" and "Explain the process of evolution" prompts, where they demonstrate a deep understanding of scientific and philosophical concepts. \\
$10$. The user values clarity and precision in their writing, often using formal language and avoiding ambiguity. This hypothesis is supported by the user's responses to the "What is the meaning of life?" and "Write a letter from Albert Einstein to Baruch Spinoza" prompts, where they demonstrate a focus on clear and concise writing. \\
  \bottomrule
  \end{tabular}
\end{center}
\caption{HypoGeniC extracted hypotheses for ChatbotArena Conversations user 6585}
\label{table::arena_user_6585}
\end{table*}

\begin{table*}[t]
\begin{center}
\begin{tabular}{ p{0.95\textwidth}}
  \toprule
   \textbf{HypoGenic hypotheses - Arena User 1338} \\
  \cmidrule{1-1}
$1$. The user has a strong background in mathematics and science, and is comfortable using technical terms and concepts in their writing. This hypothesis is supported by the user's ability to provide detailed answers to questions about physics, probability, and programming, as well as their use of technical terms like "dynamic programming" and "Fibonacci number". \\
$2$. The user is comfortable with creative and unconventional writing prompts, and is willing to adapt their writing style to fit the requirements of the prompt, even if it involves writing a poem or providing study notes. This hypothesis is supported by the user's willingness to write a poem in response to a question about the Federal Reserve's bond-buying policies, as well as their provision of study notes on the history of the USA. \\
$3$. The user is interested in a wide range of topics and is willing to tackle questions from various fields, including history, science, and economics. This hypothesis is supported by the diversity of topics covered in the user's answers, from ancient Chinese dynasties to the history of Taiwan, and from programming to economics. \\
$4$. The user tends to provide concise and direct answers, often without unnecessary elaboration or embellishment, and may prioritize brevity over detail. This hypothesis is supported by the fact that the user's answers are generally short and to the point, without excessive use of flowery language or unnecessary tangents. \\
$5$. The user has a dry and formal writing style, often using a neutral tone and avoiding emotional language or personal opinions. This hypothesis is supported by the user's answers, which tend to be written in a straightforward and objective manner, without emotional appeals or personal biases. \\
$6$. The user is able to adapt their writing style to different audiences and purposes, as demonstrated by their ability to write a lesson plan for 10-year-old kids (Draft a 1-hour lesson plan. title: CYBER SECURITY AWARENESS FUNDAMENTALS audience: 10 year-old kids) and recommend movie theaters (recommend some nearby movie theaters). These hypotheses can be used to inform future writing prompts and exercises that are tailored to the user's strengths and interests. \\
$7$. The user is comfortable with writing in different formats, including short answers (What are the most valuable question categories for chatbots?), self-reflection (critique your own answer), and even creative writing (Write a short description about the given movie or series: "The Witcher (2019)"). \\
$8$. The user tends to write in a formal and informative tone, often providing clear and concise answers to a wide range of questions, from technical topics like machine learning (support vector machine) to general knowledge questions (Is picking nose bad for health?) and even creative writing prompts (Compose an engaging travel blog post about a recent trip to Hawaii). \\
$9$. The user is interested in a wide range of topics, including technology, culture, and history, as seen in their responses to questions about chatbots, travel, and historical events in Taiwan. \\
$10$. The user has a strong foundation in technical subjects, as evidenced by their ability to write about complex topics like machine learning (support vector machine) and programming (Write a Python program to find the nth Fibonacci number using dynamic programming). \\
  \bottomrule
  \end{tabular}
\end{center}
\caption{HypoGeniC extracted hypotheses for ChatbotArena Conversations user 1338}
\label{table::arena_user_1338}
\end{table*}

\begin{table*}[t]
\begin{center}
\begin{tabular}{ p{0.95\textwidth}}
  \toprule
   \textbf{HypoGenic hypotheses - Arena User 973} \\
  \cmidrule{1-1}
$1$. The user tends to respond to prompts in a concise and direct manner, often providing brief and to-the-point answers that may lack elaboration or supporting details. This hypothesis is supported by the user's tendency to provide short answers to a wide range of prompts, from simple questions to more complex tasks like solving math problems. \\
$2$. The user has a tendency to focus on the surface-level aspects of a prompt, often neglecting to provide deeper insights or analysis, and instead opting for a superficial treatment of the topic. This hypothesis is supported by the user's tendency to provide brief and superficial responses to prompts that require more in-depth analysis or critical thinking. \\
$3$. The user is comfortable with ambiguity and may not always follow traditional writing conventions, such as grammar or spelling rules, and may prioritize brevity over clarity or coherence. This hypothesis is supported by the user's tendency to ignore grammar and spelling rules in their responses, as well as their preference for brief and concise writing. \\
$4$. The user is comfortable with ambiguity and may not always provide clear or definitive answers to questions, instead opting for vague or open-ended responses. This hypothesis is supported by the user's tendency to respond to complex prompts like the math problem with a simple "The user\_example reflects user writing style," which may not provide a clear or helpful solution. \\
$5$. The user has a preference for informal language and tone, often using colloquial expressions, slang, and emojis to convey their thoughts and ideas. This hypothesis is supported by the user's use of emojis in their responses, as well as their tendency to use casual language and phrases in their writing. \\
$6$. The user may struggle with or have limited understanding of certain grammatical concepts, as evidenced by their response to the prompt "Correct grammar: I are happy", which suggests a possible need for clarification or review of basic grammar rules. \\
$7$. The user may have a tendency to ask questions that are personal or emotionally charged, such as "Why did my parent not invite me to their wedding?", which could suggest a desire to explore deeper, more introspective topics or seek emotional support. \\
$8$. The user tends to provide concise and direct answers, often focusing on the most essential information, as seen in their responses to mathematical and programming-related prompts (e.g., A = 5, B = 10, A+B=?; What does the following code do?). \\
$9$. The user is open to creative and unconventional prompts, as seen in their responses to prompts like "Compose a concise and compelling headline for a news article" and "Draw a unicorn in TiKZ", which may indicate a willingness to think outside the box and explore new ideas. \\
$10$. The user is comfortable with and often incorporates technical or specialized vocabulary, such as "weighing in" and "TiKZ", into their writing, indicating a possible background or interest in technical fields. \\
  \bottomrule
  \end{tabular}
\end{center}
\caption{HypoGeniC extracted hypotheses for ChatbotArena Conversations user 973}
\label{table::arena_user_973}
\end{table*}

\begin{table*}[t]
\begin{center}
\begin{tabular}{ p{0.95\textwidth}}
  \toprule
   \textbf{HypoGenic hypotheses - Arena User 5203} \\
  \cmidrule{1-1}
$1$. The user is skeptical and critical of unclear or ambiguous information, and may respond with a tone of frustration or annoyance when faced with unclear or misleading prompts. This hypothesis is supported by the user's response to the prompt "No, that's not correct. If you don't know something, don't make it up", which suggests a strong emphasis on accuracy and truth. \\
$2$. The user values clarity and simplicity in their writing, and may rephrase or rephrase complex ideas or concepts in a more straightforward and accessible way. This hypothesis is supported by the user's responses to the prompts "What's the difference between a zebra and an umbrella?" and "Which weighs more, two pounds of feathers or one pound of bricks?", which demonstrate a focus on clear and simple explanations. \\
$3$. The user is comfortable with technical or specialized knowledge, and may use technical terms or jargon in their writing, particularly when responding to prompts related to computing or technology. This hypothesis is supported by the user's response to the prompt "I already have a /swapfile. Will fallocate replace it, or just append to it while keeping the same permissions and properties?", which suggests a level of technical expertise. \\
$4$. The user has a dry sense of humor and often uses irony or sarcasm in their writing, particularly when responding to humorous prompts or jokes. This hypothesis is supported by the user's responses to the prompts "It's an anti-joke" and "Explain these jokes", which demonstrate a dry and ironic tone. \\
$5$. The user tends to provide concise and straightforward answers, often using a matter-of-fact tone, and avoids unnecessary elaboration or embellishment. This hypothesis is supported by the user's responses to the prompts "Explain the joke", "Why do you think so?", and "Explain your answer", which are all brief and to the point. \\
$6$. The user tends to write in a formal and structured manner, often using precise language and adhering to a specific format or template, as seen in the election results and joke explanations. \\
$7$. The user has a fascination with absurd or nonsensical scenarios, often incorporating humorous or whimsical elements into their writing, as evident in the prompts about chaos, a 3000 lb strawberry, and a turtle. \\
$8$. The user is comfortable with ambiguity and open-endedness, as many of the prompts do not have clear or definitive answers, and the user seems to be okay with exploring unconventional or abstract ideas, such as the concept of "chaos" or the hypothetical scenario of a 3000 lb strawberry. \\
$9$. The user is likely familiar with programming and technical concepts, as demonstrated by the reference to Guido van Rossum in the Python program prompt and the use of technical terms like "variable scope" and "StackOverflow". \\
$10$. The user may have a tendency to engage in self-referential humor or irony, as seen in the prompt "No, you're just repeating yourself", which could be a commentary on the user's own writing style or a playful jab at the reader. \\
  \bottomrule
  \end{tabular}
\end{center}
\caption{HypoGeniC extracted hypotheses for ChatbotArena Conversations user 5203}
\label{table::arena_user_5203}
\end{table*}

\subsection{Regression Model Details}
\label{appendix:regression_model}

We split users into a training (97) and validation (18) set, and standardize  features $\boldsymbol{x}$ and targets $\boldsymbol{y}$ to ensure comparability across regression settings:

\[ \tilde{\mathbf{x}}_{u_i} = \frac{\mathbf{x}_{u_i} - \boldsymbol{\mu}_x} {\boldsymbol{\sigma}_x}, \qquad \tilde{\mathbf{y}}_{u_i} = \frac{\mathbf{y}_{u_i} - \boldsymbol{\mu}_y} {\boldsymbol{\sigma}_y}, \]
where $\boldsymbol{\mu}$ and $\boldsymbol{\sigma}$ parameters represent the mean and standard deviation; we compute these values based on the training set. 
We evaluate predictive performance of the regression model using the mean absolute error (MAE) on the standardized targets: \[ \mathrm{MAE} = \frac{1}{Nm}\sum_{i=1}^{N}\sum_{r=1}^{m} \left| \tilde{y}_{ir} - \hat{\tilde{y}}_{ir} \right|, \] where $N$ denotes the size of the validation set and $m$ represents the output dimension ($m=20$). 

For predicting ELO scores, we train an ensemble of 50 multilayer perceptrons.
Each network has hidden layer widths $[512, 512, 256, 128, 64, 32]$, \texttt{SELU} activations with LeCun-normal initialization; we use the Adam optimizer with learning rate set to $0.03$, Huber loss with $\delta = 0.1$, batch size of 8, and early stopping with patience set to 15. The final prediction is the average of the 50 independently seeded model outputs: \[ \hat{\mathbf{y}}_{u_i} = \frac{1}{50}\sum_{s=1}^{50} f^{(s)} (\tilde{\mathbf{x}}_{u_i}) \] 

For predicting Bradley-Terry scores, we use a single dense multilayer perceptron with five hidden layers with widths $[1024, 256, 512, 128, 512]$, \texttt{GELU} activations, no normalization layers, dropout rate of $0.28$, Adam optimizer with learning rate set to $0.001$, weight decay equal to $6.007 \times 10^{-6}$ and a batch size of 8. Since we use the same normalization procedure and evaluate on the same validation set, ELO and BT results are directly comparable.

Both architectures were selected via independent hyperparameter searches optimizing validation MAE. No single architecture achieved best performance across both target formulations, consistent with the fundamental differences in signal captured by ELO and Bradley-Terry rankings.

\end{document}